%% file: noisebench.tex
\pdfoutput=1
\documentclass[11pt]{article}

\usepackage[preprint]{acl}

\usepackage{times}
\usepackage{latexsym}

\usepackage[T1]{fontenc}
\usepackage[utf8]{inputenc}

\usepackage{microtype}

\usepackage{inconsolata}

\usepackage[disable]{todonotes}
\usepackage{booktabs}
\usepackage{subcaption}
\usepackage{graphicx}

\usepackage{makecell}

\def\benchmark{\textsc{NoiseBench}}
\def\noiseclean{Clean}
\def\noiseexpert{Expert}
\def\noisecrowd{Crowd}
\def\noisecrowdbest{Crowd++}
\def\noisedistant{Distant}
\def\noiseweak{Weak}
\def\noisellm{LLM}

\title{\benchmark: Benchmarking the Impact of Real Label Noise \\ on Named Entity Recognition}  

\author{Elena Merdjanovska\textsuperscript{1,2}, Ansar Aynetdinov\textsuperscript{1}, Alan Akbik\textsuperscript{1,2} \\ 
 \textsuperscript{1}Humboldt-Universität zu Berlin \\ \textsuperscript{2}Science of Intelligence \\
 \texttt{\{elena.merdjanovska, aynetdia, alan.akbik\}@hu-berlin.de}}

\begin{document}
\maketitle
\begin{abstract}

Available training data for named entity recognition (NER) often contains a significant percentage of incorrect labels for entity types and entity boundaries. Such \textit{label noise} poses challenges for supervised learning and may significantly deteriorate model quality. To address this, prior work proposed various \textit{noise-robust learning} approaches capable of learning from data with partially incorrect labels. These approaches are typically evaluated using \textit{simulated noise} where the labels in a clean dataset are automatically corrupted. However, as we show in this paper, this leads to unrealistic noise that is far easier to handle than real noise caused by human error or semi-automatic annotation. To enable the study of the impact of various types of real noise, we introduce \benchmark, an NER benchmark consisting of clean training data corrupted with 6 types of real noise, including expert errors, crowdsourcing errors, automatic annotation errors and LLM errors. We present an analysis that shows that real noise is significantly more challenging than simulated noise, and show that current state-of-the-art models for noise-robust learning fall far short of their achievable upper bound. We release \benchmark~for both English and German to the research community\footnote{\url{https://github.com/elenamer/NoiseBench}}.
\end{abstract}

\section{Introduction}

Named entity recognition (NER) is the task of detecting and classifying named entities in text, such as the names of organizations or locations. Current state-of-the-art approaches for NER still require supervision in the form of labeled training data \cite{zaratiana2023gliner}, i.e.~sentences in which named entities are marked and assigned their correct type. 
However, prior work found that available datasets for NER and other supervised tasks are affected by \textit{label noise}, meaning that a certain percentage of entity labels are incorrect. For instance, the common NER dataset CoNLL-03~\cite{tjong-kim-sang-de-meulder-2003-introduction} was estimated in various studies to have noise shares of between 5 and 7\%~\cite{Wang2019CrossWeighTN,Reiss2020IdentifyingIL,ruecker2023clean}. Other NER datasets have also been found to contain a share of incorrect labels, with OntoNotes4 estimated around 8\% and WNUT-17 around 18\% \cite{Wang2019CrossWeighTN,huang-etal-2021-named}. 

% \begin{figure}[t!] %[htpb]
%     \centering
%     	\includegraphics[width=\linewidth]{figures/example_CoNLL.pdf}
%     \vspace{-6mm}
%     \caption{\small An example snippet from the CoNLL-03 named entity recognition dataset, with the correct label (top row), real label noise (middle row) and simulated noise (bottom row). The crowdsourced label only annotates the portion "Royal College of Physicians" as an entity of type ORG (organization), a plausible annotation that is however incorrect by the guidelines of the dataset. The commonly used class-dependent simulation approach for label noise is much less plausible. 
%     }
%     \vspace{-2mm}
%     \label{fig:example-sentence}
% \end{figure}

\input{figures/text_example_figure}

Label noise introduces inconsistencies during training, which %complicates the training signal with inconsistencies, and 
may significantly deteriorate model quality~\cite{zhang2021generalization}. To address this issue, prior work proposed approaches for \textit{noise-robust learning} aimed at mitigating the negative effects of the noisy training signal~\cite{song2022learning}. However, the evaluation of these approaches has two main limitations.
\todo[]{introduction seems a bit long?}

\noindent 
\textbf{Limitation 1: Simulated noise is too easy.}  %Limitation 1: evaluating with simulated noise.
Most current research in noise-robust learning relies on experiments with \textit{simulated label noise}~\cite{tanzer-etal-2022-memorisation, klie2023annotation}. While this allows for evaluation in a controlled setting, it has been shown that simulated noise, even though it can model noise well to some extent, is much easier for deep learning models to disregard than real label noise~\cite{pmlr-v119-jiang20c,Zhu2022IsBR}. %maybe another sentence why simulated noise is bad, and why developing better simulation models is bad

Refer to Figure~\ref{fig:example_sentence_figure} for an illustrative comparison between real and simulated noise for three example sentences, including different types of errors that occur in NER datasets. These examples demonstrate that simulated noise can introduce similar errors as real noise, however the choice of spans to mislabel is random and as a result often less plausible. This means that an approach shown to be robust to simulated noise may not in fact be robust to real noise in practice.

% The real noise (middle row) contains a plausible but incorrect annotation. The simulated noise (bottom row) on the other hand is the result of (informed) random sampling, which effectively models noise to a certain degree, however it sometimes introduces implausible entity boundaries and types. As there is no underlying pattern to this example of simulated noise, it poses fewer challenges. 
% This means that an approach shown to be robust to simulated noise may not in fact be robust to real noise in practice.

\noindent 
\textbf{Limitation 2: Distinct types of real noise.} Additionally, there exist many possible sources of "real" noise. For instance, expert labelers may make different mistakes than crowd workers~\cite{frenay2014survey}. Next to human labeling, there are widely-used automatic approaches to create NER-labeled datasets such as distant supervision from a knowledge base~\cite{mintz2009distant,hedderich2021analysing} and weak supervision using rules~\cite{zhang2021wrench}. Lastly, current research investigates the use of LLMs to label datasets~\cite{golde2023fabricator,Wang2023InstructUIEMI}.
\todo[]{(option1) choose one image of Figure 2, put the others in appendix}

We postulate that these types of real noise differ in their characteristics, meaning that a noise-robust learning approach shown to perform well on one type of noise may not perform well on another. For this reason, we argue there is a need for evaluating noise-robustness across multiple label noise types.  

\noindent 
\textbf{Contributions.} With this paper, we present \benchmark, a new benchmark for measuring the impact of label noise in the training data on the prediction quality of trained NER models. In more detail, our contributions are: 
\vspace{-2mm}
\begin{itemize}
\itemsep0em 
    \item We construct a noisy training dataset in 7 different variants, where each noisy variant contains the same sentences and is affected by one class of real errors, spanning errors made by experts, crowd workers, distant supervision, weak supervision and teacher LLMs. 
    \item We present a set of experiments that empirically show that real noise from \benchmark{} is significantly more difficult for current approaches. We further find that during training, real noise is memorized immediately, whereas memorization of simulated noise is delayed.
    \item We comparatively evaluate current state-of-the-art approaches for noise-robust learning on \benchmark{}, and experimentally establish upper bounds. 
 %   \item We verify our findings for German, on a new benchmark with 2 noisy training variants.
\end{itemize}
\vspace{-1mm}
Our analysis finds that no single current approach works best for all types of real noise, and that all current approaches fall far short of their theoretical upper bound. To enable the research community to leverage our benchmark in their evaluations, we publicly release all data and implementation.
%\footnote{Link to be made public upon acceptance.}

\todo[inline]{strengthen contributions. also make it clear that simulating noise is what everyone else is doing and creating an ideal NER noise model is also challenging.}

\section{\benchmark{}}
\label{benchmark}
Our benchmark is derived from a subset of the classic CoNLL-03 dataset for NER in English, annotated with entities belonging to four classes. We chose this dataset since it has been extensively studied in the field, allowing us to integrate various prior works. We derive a similar benchmark for NER in German, in Section \ref{sec:ablation_german} and Appendix \ref{appendix-german-benchmark}.

\benchmark{} consists of the following parts: (1)~A noise-free test split to evaluate trained models. (2)~Seven variants of the training split, where six are annotated with different types of noise and one is without noise. 
Table \ref{conll_noise_shares_table} presents the quality of the six noisy variants w.r.t. the noise-free dataset.
%The quality of the six noisy variants, in reference to the noise-free dataset, is presented in Table \ref{conll_noise_shares_table}.

The training split contains 5,885 sentences from 400 documents, covering 9,685 entity mentions. The test split 
contains 3,427 sentences from 231 documents, covering 5,725 entity mentions. 
%Approaches should be separately trained over each noise split, and evaluated on the clean test data. 

%One of the variants is a clean version without any noise to measure the upper bound (best achievable score if trained on clean data). 

%We compile 6 noisy label sets from different sources of noise, presented in Table \ref{conll_noise_shares_table}. The table shows both token-level metrics (significant if we want to discuss NER as a noisy classification problem), sentence-level metrics (significant if we want to include only clean sentences in training) and entity-level metrics (metrics of interest in NER). All metrics are expressed as percentages. For this task, we define the noise level in terms of the entity-level F1 score, as $100 - \%F1$.
%In this section, we first discuss different sources of real label noise and how we derive corresponding noise split for \benchmark{}. We then present dataset statistics and discuss methods for simulating label noise.

\input{tables/noisebench_overview_shares}

\subsection{Types of Noise} 

In the following, we discuss each training split and the type of noise it models.

\subsubsection{Noise-Free Data}

Our benchmark requires two splits without any label noise: A clean test split to evaluate models trained on noisy training data, and a \textbf{\noiseclean} training split to measure the upper bound performance. 
%This allows us to compare models trained on noisy training splits against the F1 score of models trained on the clean training split. 

%\textbf{Derivation of "Clean" split.}
Since the original annotations of CoNLL-03 have been shown to be noisy~\cite{Wang2019CrossWeighTN, Reiss2020IdentifyingIL}, we use the labels of \textsc{CleanCoNLL}~\citep{ruecker2023clean}, a recently released resource in which 7\% of all original annotations were semi-automatically relabeled. In their evaluation, \citet{ruecker2023clean} find their resulting dataset to be of very high quality and largely improved consistency. %They also showed that state-of-the-art NER methods reach significantly higher F1 scores on their cleaned version of CoNLL-03. 
The \textbf{Clean Test} split in our benchmark is the standard CoNLL-03 test split, with the \textsc{CleanCoNLL} labels.

\subsubsection{Expert Errors}

Most machine learning datasets are created using manual annotation by domain experts that provide high-quality labels. However, errors have been found to occur even in expert annotation, affecting even well-known benchmarks, though usually with relatively low noise shares of under 10\%~\cite{northcutt2021pervasive,song2022learning}. 
To represent such noise, our benchmark includes a variant of the train split called \textbf{\noiseexpert}, which contains the original CoNLL-03 annotations. As Table~\ref{conll_noise_shares_table} shows, this split has a noise share of 5.5\% and is thus the split with lowest noise.

%, as an example of a widely-used benchmark, which has later been identified to include errors. 

%Human annotation errors can happen for different reasons, including accidents, insufficient information and inter-rater variability \cite{frenay2014survey} and they can occur even with expert annotators \cite{song2022learning}. 

\subsubsection{Crowdsourcing Errors}

Crowdsourcing is a less costly alternative to expert annotation, but also more prone to 
annotation errors %, which can happen for different reasons , including accidents, insufficient information and inter-rater variability 
\citep{frenay2014survey}. In order to create noisy variants of the train set representing real-world human errors, we utilize the crowdsourced labels by \citet{rodrigues2014sequence}. This study involves 47 crowd workers labelling a subset of the English CoNLL-03 dataset, of around 400 news articles. %, with each crowd worker labeling a subset of these articles. 
They released their dataset and all annotations produced by each crowd worker. We selected only the sentences where the tokenization matched the Clean variant, resulting in 5,885 sentences.  

%Comparing the annotation quality of the crowd workers to our \noiseclean~split, we measure large discrepancies between each worker, from the "best" worker labeling at 0.85 F1 score to the worst labeling only at 0.17 F1. As such discrepancies are expected, much crowdsourcing research focuses on finding efficient ways to filter and aggregate annotations produced by the crowd.

We include two noisy training splits based on crowd annotations into our benchmark: (1)~In the first, \textbf{\noisecrowd}, we do a simple majority vote over all annotations provided for each token, i.e.~the baseline method for aggregating crowdsourced annotations. (2)~In the second, \textbf{\noisecrowdbest}, we use an oracle version of the majority vote, selected by either taking the correct label if it is provided by any of the annotators or, in the absence of a correct label, by choosing the label with the majority of votes. This version represents the upper bound of crowdsourced labels given a perfect label aggregation method. As Table~\ref{conll_noise_shares_table} shows, the noise share of \textbf{\noisecrowd} (36.6\%) is considerably higher than \textbf{\noisecrowdbest} (15.3\%).

\subsubsection{Distant Supervision}

One approach for labeling data without human participation is \textit{distant supervision}~\cite{mintz2009distant}, where entity mentions in target datasets are matched to entity types in knowledge bases (KBs). %The matching can be achieved by simple string matching, use of regular expressions or heuristics.

% \todo[inline]{A few more details on the process. Longest entity names are matches first? Error sources: Incompleteness in the KB? Wrong matching?}

We include a \textbf{\noisedistant~}noisy training variant in our benchmark, adapted from the annotations by \citet{liang2020bond}\footnote{\label{license}Available under Apache 2.0 license} that use the Wikidata corpus and gazetteers collected from multiple online sources as external knowledge bases. After initial POS tagging%with an NLTK tagger \cite{loper2002nltk}
, the unlabeled sentences were matched with the knowledge bases. This process results in incomplete annotations due to limited coverage over entity types of KBs. %As a result, the KB matching yields a large amount of instances labeled with class "O". 
This explains the rather high number of missing entities %total number of entities
and the overall noise level (31.3\%) of the \textbf{Distant} training variant, as shown in Table~\ref{conll_noise_shares_table}. 

\subsubsection{Weak Supervision}

Another approach aimed at reducing manual annotation efforts is weak supervision. Here, labels are obtained using a number of ``weak'' supervision sources, such as heuristics or expression-based rules. Each weak source is typically specialized to detect only a subset of the correct labels. %Disagreements between the resulting weak labels can be resolved by voting schemata or more complex unsupervised methods.

We use the labels from the approach by \citet{lison-etal-2020-named}\textsuperscript{\ref{license}} to create our \textbf{\noiseweak}~label set. This covers 16 weak labeling sources \cite{zhang2021wrench}, including heuristics, gazetteers %(Crunchbase, Geonames, Wikidata)
and predictions of NER models trained on other corpora% (Broad Twitter Corpus and Ontonotes5.0)
. An example heuristic is detecting PER (person) entities using a pre-defined list of %English 
first names.

We aggregate the weak label sets with simple majority voting. We apply majority vote on every token with at least one entity label assigned to it, following \citet{zhang2021wrench}. Due to the large number of labelling sources, majority voting yields a large number of entities, as shown in Table \ref{conll_noise_shares_table}, including many false positives. As a result, the \textbf{\noiseweak~}label set has a high noise share of 40.4\%.

\subsubsection{LLM Teacher Models}

%Different areas in NLP have witnessed remarkable advances with \textbf{large language models (LLMs)}. 

Our benchmark includes a noisy variant of the train split annotated by an LLM. This follows recent efforts that use LLMs for dataset generation~\cite{Wang2023InstructUIEMI}. Here, the main idea is to pass a description of the annotation task and target classes to an %instruction-tuned 
LLM, and provide sentences to label. LLMs are able to generate high quality labels for some tasks (e.g. sentiment classification) while for others (e.g.~NER and question type categorization) the resulting labels are very noisy \cite{golde2023fabricator}. 

We created the \textbf{\noisellm~}variant using the Fabricator toolkit \cite{golde2023fabricator} by prompting GPT3.5 for named entities in our training dataset. To use LLM outputs for annotation of NER datasets, a certain output format is required. To achieve this, we provide one example with the correct output format in each prompt. This example is the same for each sentence we wish to annotate, which we refer to as a static one-shot setting. The example sentence was selected from the remainder of the CoNLL-03 training split, which consists of all sentences not included in our benchmark.

As Table~\ref{conll_noise_shares_table} shows, the \noisellm~label set results in the highest noise share of 45.6\%. This is mainly due to the large number of nouns incorrectly identified as entity mentions, which also makes this the label set with the largest number of entity annotations out of the variants in \benchmark.

% and a teacher LLM approach (\textbf{\noisellm} label set), where we used GPT3.5 in a static one-shot setting. As shown in \ref{conll_noise_shares_table}, these approaches result in higher noise levels, when compared to human annotation noise.
\subsection{Statistics}
\vspace{-1mm}

An overview of \benchmark~is given in Table \ref{conll_noise_shares_table}. 
% With the metrics shown, we assess the performance between the \noiseclean~training split and the noisy variants. 
The table shows token-level F1 score %(\textit{F1\scriptsize{token}}\normalsize)%(significant if we want to discuss NER as a token classification problem)
%, sentence-level %(significant if we want to include only clean sentences in training) 
 and entity-level F1 score %(\textit{F1\scriptsize{entity}}\normalsize), %(metrics of interest in NER). 
expressed as percentages. We define the noise level (\textit{\%Noise}) in terms of the entity-level F1 score, as $100 - \%F1$.
The noise levels of the noisy splits range from 5.5 to 45.6 percent.

 %the table shows the total number of entities, the number of correct entities, as well as the share of different error types. 
The table also shows the share of different error types.
The errors are categorized into 4 main categories: \textbf{missing} mentions, \textbf{non-entity} mentions (false positives), incorrect entity \textbf{type} (where the boundary is correct, but type incorrect) 
and \textbf{partial} matches. Partial matches are special cases where the type is correct, but the mention boundary is only partially correct. Refer to Figure~\ref{fig:example_sentence_figure} for examples.

% Notably, the \noisecrowd~label set results in a noise share comparable to the \noisedistant~set, both in terms of F1 scores and of total number of entities annotated. % why is it important to highlight? is it really that close? Weak and Crowd are comparable as well by that logic

We observe that the \noisecrowdbest, \noisecrowd~and \noisedistant~label sets have a lower total number of entity annotations than the \noiseclean~dataset, and the largest portion of errors are missing mentions. Conversely, the \noiseweak~and \noisellm~label sets have more annotations than the \noiseclean~dataset, and most of the errors are either an incorrect mention or incorrect type. Most of the errors in the \noiseexpert~label set are due to incorrect type. Regarding the number of partial matches, for almost all noise types, they make up between 10\% and 15\% of all errors. 

%% To remove: This, in combination with their higher precision and lower recall scores, signifies that there is a large number of missing entity mentions in these label sets, however the entity mentions that are in fact included in the labels are largely correct. Conversely, the annotations in the \noisellm~and \noiseweak~label sets have low recall scores, even though they output more entities than present in the \noiseclean~dataset, meaning the abundance of annotations still misses or misclassifies a large number of the correct entities. Moreover, \noisellm, \noiseweak, \noisecrowd~and \noisedistant~ are all characterized with comparable recall scores, meaning they have a similar number of missing target entity mentions.

%% To remove: When looking at the overall share of correct sentences in each noisy variant, we can see that, as expected, it decreases with increasing noise levels. As a result, around 50\% of the sentences in the \noiseweak~variant and only 38\% of the sentences in the \noisellm~variant are correctly annotated. 

% next iteration: how many of these sentences have entities (how many entities in total does the clean subset include)?

\section{Comparing Real and Simulated Noise}
\vspace{-1mm}

%The main goal of our experiments is to evaluate noisy label learning approaches on the datasets in our benchmark. 

% In the first two experiments we use \benchmark~to investigate the impact of each type of noisy labels on model performance. Furthermore, we compare training under real and simulated label noise in order to highlight the importance of evaluating noise-robust learning approaches on real label sets instead of simulated ones.

We first use \benchmark~to investigate how real label noise affects NER model performance in comparison to simulated noise. For this, we conduct two experiments: the first one addresses the impact of each type of training noise on the clean test set performance, and the second one compares training dynamics under real and simulated label noise to highlight the differences in noise memorization.

% for both NER and sentiment classification.

%Experiments are concerned with the noisy label learning scenario, which includes training on a noisy variant of the training set, and maximizing the score on the clean test set. 

% , however our goal was to evaluate these approaches in a setting, where the labels for all available data are noisy. 

% \input{tables/Exp1_real_noise_extended_scores}

% \input{tables/Exp1_simulated_noise_extended_scores}

\subsection{Noise Simulation Methods}

We consider two noise simulation methods, namely the simple \textit{uniform noise} used in most prior work and a more involved \textit{oracle class-dependent noise} method that we design to mirror each noisy variant in \benchmark{}.

% more involved method for simulating realistic noise that combines various proposals from prior work. 

\noindent 
\textbf{Uniform noise.} %The most common simulation approaches use either uniform or class-dependent noise\cite{frenay2014survey}. 
Uniform noise corrupts samples into any other label with a uniform probability distribution, given a target noise share. Studies investigating simulated noise in the NER task commonly rely on variants of this method \cite{mayhew-etal-2019-named,tanzer-etal-2022-memorisation}. 

%This matrix can be defined from prior knowledge of the problem or by analyzing the mislabeling frequencies in realistic noisy label sets.

%In the context of named entity recognition, we could for example discern that PER (person) and ORG (organization) entities are more likely to be mislabeled than PER and LOC (location). 

% This differs from other studies investigating simulated noise for NER on CoNLL-03, since variants of uniform noise are commonly used in \citet{mayhew-etal-2019-named,tanzer-etal-2022-memorisation}, even though the class-dependent model is more realistic. Since label noise models are mainly defined for classification tasks, we observe NER as a token classification task. 

%Since label noise models are mainly defined for classification tasks, we observe NER as a token classification task. 

\noindent 
\textbf{Oracle class-dependent noise.} 
Class-dependent noise is based on the knowledge that some pairs of classes are more likely to be mislabeled than others. It is defined by a noise transition matrix, which contains the mislabeling probabilities between all pairs of classes \cite{hedderich2021analysing}. We design an oracle version of class-dependent noise, where the per-class mislabeling probabilities of real noise are known. This allows us to investigate class-dependent noise in an ideal case, where it is able to mirror real noise closely, even though this is not possible in practice. This method mirrors real noise by utilizing the token-level mislabeling frequencies as probabilities to form a noise transition matrix.

Using each noise simulation method, we created 6 label sets, corresponding to each noise level in \benchmark{}. It should be noted that the simulated labels replicate the token-level F1 scores of the real noisy labels, however the entity-level F1 and sentence-level accuracy can deviate. 

\subsection{Experimental Setup}
%\textbf{Experimental setup.}
% In this experiment we focus on the NER task. 
% We use a baseline approach, which involves fine-tuning a pre-trained transformer-based model. 

In both experiments, we train a baseline approach for NER on each noisy variant of the training split, as well as on the additional simulated noise. 
% Experiment 1 compares the impact of different noise types on test set performance, and Experiment 2 evaluates their impact on the training curves and investigates the memorization effect.

\noindent 
\textbf{Validation splits.} We evaluate the setting in which all available data to train a model is noisy, including the validation set. To obtain noisy validation sets for each of our 7 dataset variants, we split the noisy datasets into training and validation sets. All sentences from 66 news documents from 1996-08-24 comprise the validation set, which is around 17\% of all sentences, and are left out from model training and used for hyperparameter tuning. 

\label{sec3:baseline}
\noindent
\textbf{Baseline.} For NER, as a baseline approach, we fine-tune an \texttt{xlm-roberta-large} transformer using the FLERT approach \cite{schweter2021flert}. It improves upon the regular fine-tuning setup by considering document-level features of a sentence to be tagged. %: for each sentence, 64 subtokens of left and right context (surrounding sentences in the document) are added. 
We use a learning rate of 5e-6 and a batch size of 32, for a fixed number of 10 epochs, without early stopping% linear scheduler with warmup.
. These parameters were obtained according to the performance on a noisy validation set, keeping in mind that larger batch sizes are more robust to noise \cite{rolnick2017deep}. 
We use entity-level micro F1-score.

\subsection{Experiment 1: Impact of Label Noise on Test Performance}
\label{Exp1}

% In the first experiment we compare the 6 variants of real label noise in our benchmark with simulated noisy labels in terms of test set performance. %With this we aim to investigate the impact of each label type on model performance and highlight the importance of evaluating noisy label learning approaches on real label sets instead of simulated ones. 

In the first experiment, we compare how the clean test set performance is impacted by the 6 types of real label noise when present in the training set. In addition, we provide the same comparison for corresponding simulated noisy label sets.

% \vspace{-2mm}

%For this experiment we train the baseline model for a fixed number of 10 epochs.

% \todo[]{add simulated noise levels to table 3 - either only entity-F1 or entity-F1, token F1 and num. entities}

\subsubsection{Results and Discussion}

The results for uniform noise are shown in Appendix \ref{sec:appendix-simulated-noise-extended}. We initially established that uniform noise is less challenging for the model, so in the results for Experiments 1 and 2 we chose to focus solely on oracle class-dependent noise.

The main results from Experiment 1 for oracle class-dependent noise are shown in Table \ref{conll_baselines_real_simulated}. Following are our main observations.

\noindent
\textbf{Label noise degrades performance.} When we compare the test F1 scores of the real noisy variants with the average score of 93.99 achieved when training on the \noiseclean~variant, we see that model performance is affected by each noise type.
%In all cases, the noisy training variants yield lower test scores, even when training with the low-noise \noiseexpert~variant. 
As the noise levels increase, this impact becomes more pronounced, showing that the baseline model lacks robustness to real noise. 
Comparing the test F1 scores of the simulated noisy variants, we can see that noise of 5.9\% in the training set %does not hurt model performance and
results in a score comparable to training on the \noiseclean~variant. However, as simulated noise levels increase, the noise does degrade test set scores.

\input{tables/simulated_vs_real_noise_shortened}

\noindent
\textbf{Real noise is more difficult.} Furthermore, when we compare the real noisy label sets with their equivalent simulated noisy variants, we can observe that the simulated training variants show a score of around 2.5 percentage points higher on average than the real label sets. %This happens in the case of most label sets, excluding the weak supervision labels, where the
This shows that for predictive NER models, real noise is more difficult to overcome than simulated noise. In other words, models are more likely to overfit to real noisy labels, rather than simulated ones.

\input{figures/seen_vs_unseen_plots}

\noindent
\textbf{Models generalize to unseen entities well.}
Figure \ref{fig:seen_vs_unseen} shows F1 scores for seen and unseen entities separately, further distinguishing seen entities by whether their label in the training set was clean or noise-corrupted. 
In Figure \ref{fig:seen_vs_unseen_real} we see that for the \noiseexpert~and \noisecrowdbest~noise types, the score on the seen (clean) and the unseen entities is comparable, which indicates the model has the ability to generalize to unseen entities well. As for the remaining training splits with noise levels of over 30\%, noise also affects the performance on unseen entities.

\noindent
\textbf{Simulated noisy patterns are disregarded.} For all real noise types, the score on the seen (noisy) entities is low. With simulated noise however, in Figure \ref{fig:seen_vs_unseen_simulated} we see that for \noiseexpert~and \noisecrowdbest, the score on the seen-noisy entities and seen-clean entities is close. This means that at low noise levels, the models are able to disregard simulated noisy patterns and predict the same entities correctly when they appear in the test set.

\input{figures/memorization_plots_grid}

\subsection{Experiment 2: Memorization of Noise}
\label{memorization}

%The second experiment investigates the memorization of label noise. We compare this effect for real and simulated noise. 

%\subsubsection{Experimental Setup}
%\textbf{Experimental setup.} 
% We use the technique to simulate label noise described in the first experiment in Section \ref{Exp1}. We again evaluate the baseline models when fine-tuning on the 6 noisy variants in our benchmark, as well as on the 6 simulated noisy label sets with comparable levels. 

% Unlike Experiment 1....

Prior analysis has found that there are distinct phases of learning when training a model on data with label noise~\cite{arpit2017closer}. This has been referred to as a \textit{generalization phase}, where models learn patterns that generalize well to clean data, followed by a \textit{memorization phase}, where models overfit to the label noise and deteriorate in prediction quality~\cite{tanzer-etal-2022-memorisation}.  

%This has been refered to  distinct phases of training with noisy labels: a generalization phase, a settling phase (during which the training score does not change) and a memorization phase.  
To investigate this phenomenon for real and simulated noise, we extend the training stage to 100 epochs. At the end of each epoch, we measure the F1 score of the model on both the noisy training split it is being trained on, and separately on the clean training split. The difference between these two scores allows us to measure %generalization and 
memorization.

%\citet{tanzer-etal-2022-memorisation} have identified that noise memorization in transformer-based models occurs with a certain delay %of around 4 training epochs, after which the models start overfitting to training label noise. 

% In the second experiment we are interested in the training dynamics during training epochs and aim to observe the memorization process, therefore extending the training to 100 epochs. 

\subsubsection{Results and Discussion}

%\textbf{Results.}

%\todo[]{one paper uses Measurable performance for the blue lines and True performance for the orange lines, maybe think about using that in our text}

In Figure \ref{fig:memorization_figure} we show training curves from training with real and simulated variants of \benchmark~ for 3 noise types: \noiseexpert, \noisecrowdbest~and \noisedistant. We plot two scores: the F1-score on the respective noisy variant of the training set, and the F1 score on the \noiseclean~variant of the training set. 
%, also highlighting the 10th epoch, where training stops in our baseline approach. 
In all training curves, we can observe the memorization effect, with each model perfectly fitting the noisy data by the end of training and reaching an F1 score close to 1.

% next iteration: \todo[]{think about discussing 'settling' phase, we do see it in the plots, but mentioning it doesn't add much}

\noindent
\textbf{Delayed memorization of simulated noise.}  However, we note that with simulated noise (see Figure \ref{fig:simulated_original}, \ref{fig:simulated_mv_oracle}, \ref{fig:simulated_bond}) this happens much later in the training process than with real noise. In addition, the training curves of simulated noise show a stage during the early epochs where the score on the clean labels is consistently higher than the score on the noisy labels. This confirms previous findings that the model is able to learn general patterns first, before starting to memorize the noise. 

\noindent
\textbf{Immediate memorization of real noise.} With real noise, this does not happen and the model starts fitting the noisy labels from the beginning (see Figure \ref{fig:real_original}, \ref{fig:real_mv_oracle}, \ref{fig:real_bond}). As a result, the score on the clean labels is consistently lower than the score on the noisy labels, throughout the training run\footnote{We confirm this finding for German in Appendix \ref{appendix-memorization-german}.}.

Our experiments find that real noise does not display distinct generalization/memorization phases during training, and rather immediately begins with memorization\footnote{We confirm this finding for a smaller model, as well as a randomly initialized model in Appendix \ref{sec:appendix-memorization}.}. This makes intuitive sense, as real noise has underlying patterns that may be extracted during learning. This lends further evidence to the increased challenges and the need to evaluate noise-robust learning with real noise.

% Maybe because there are general patterns to be learned in real noise, while no patterns in simulated?

% We can observe that in the cases with real noisy labels (\ref{fig:real_original},\ref{fig:real_mv} and \ref{fig:real_bond}), the F1 score on the noisy training set is higher than the score on the clean set through all epochs. This indicates that the model is fitting the noisy labels and does not generalize well to the clean labels. When simulated noise is present (\ref{fig:simulated_original},\ref{fig:simulated_mv} and \ref{fig:simulated_bond}), we can observe the opposite effect, namely, the F1 score on the clean training set is higher than the score on the noisy set through all epochs. This indicates that the model memorizes this type of noise much less and is able to maintain generalization abilities.  %Pre-trained language models are inherently robust to some types of noise \cite{Zhu2022IsBR}, which means that in some cases high scores on the clean training set can be expected. 

% \input{figures/memorization_plots_grid}

% TAKEAWAY: 
% \begin{itemize}
% \item Realistic noise has immediate memorization
% \item Simulated noise has delayed memorization
% \end{itemize}

\section{Evaluating Noise-Robust Learning}

Having established the difficulty of real noise, we now use \benchmark{} to perform a comparative evaluation of widely-used noise-robust learning approaches. Our goal is to determine their effectiveness in the presence of real label noise, and to establish upper bounds of what noise-robust learning could ideally achieve.

%Lastly, we aim to evaluate widely used noisy label learning approaches on the different types of real noise we have considered so far, in order to determine their effectiveness in the presence of real label noise. 
% To investigate this problem, we evaluate various methods on both NER and sentence classification. 

\subsection{Compared Approaches}

We surveyed current state-of-the-art methods for noise-robust NER and found that many approaches rely on the same underlying ideas for handling label noise. In the following, we group approaches by the underlying idea, select a state-of-the-art representative for each group and, if possible, derive an upper bound method for each group. For more details about the implementation of compared approaches refer to Appendix \ref{appendix:implementation-details}.

\subsubsection{Learning from a Clean Subset}
\label{clean_subset}
The first family of approaches relies on utilizing the subset of each noisy dataset in which all labels are correct. One type of these approaches filters out all likely incorrect annotations and learns only from a clean subset. Another type derives confidence weights for each sample so that annotations judged to be of higher quality feature more during training.

As a representative of the former type of approaches targeting clean subsets of noisy datasets, we chose Confident Learning~\cite{northcutt2021confident}, while the latter type is represented by CrossWeigh~\cite{Wang2019CrossWeighTN} and Learn-To-Reweight (L2R) \cite{ren18l2rw}.

%\textbf{Confident learning.} 
 %It works in combination with any model which outputs probability scores. Given the predicted probabilities and the noisy labels, it estimates a joint distribution of noisy and true labels. Using this distribution, for each class, the samples with the highest probabilities that belong to a class other than the observed one are flagged as errors.
\noindent
\textbf{Upper bound: Oracle subset.} To obtain an upper bound for this family of approaches, we use an oracle to select the subset of clean sentences from each of the noisy training splits in \benchmark{}. We then use the baseline fine-tuning approach only on this subset, illustrating a best-case scenario.

%This upper bounds represents the best-case

\subsubsection{Delaying Memorization}

Another family of noise-robust learning approaches seeks to leverage the two phases of learning (generalization and memorization) we discussed in Section~\ref{memorization}. They seek to either draw our the generalization phase or cease training before memorization begins. While our experiments indicate that these two phases do not exist for real noise, we nevertheless include this family of approaches in our evaluation since they are widely used. % and to gain additional insights into their behavior on real noise. 
As representative of this class of approaches, we chose co-regularization \citep{zhou2021learning}.

%\textbf{Co-regularization.} , which exploits the finding that noisy labels often have delayed learning curves by jointly optimizing multiple classifiers on the Kullback–Leibler divergence among predicted probability distributions in addition to the task-specific loss of each individual classifier. 

\noindent
\textbf{Upper bound: Oracle stopping.} To obtain an upper bound for this family of approaches, we use a simple stopping criterion based on the score on %the true generalization using 
the clean test set at the end of each epoch. We use the epoch of best generalization to report the final score. This simulates an ideal stopping. %(albeit at the granularity of full epochs). 

\input{tables/conll_results}
% \subsubsection{Sample Reweighting}

% The third family of approaches modifies the loss calculation by assigning lower weights to potentially noisy samples. We chose CrossWeigh \citet{Wang2019CrossWeighTN} as a representative of this family of approaches.

% \textbf{Upper bound: Oracle weights.} We use the loss modification from CrossWeigh, but assign optimal weights to each sample: 1.0 for the tokens in clean sentences and 0.1 for the tokens in noisy sentences. (If we use weight=0, then this is the same as oracle subset) - what is the actual minimum weight in CrossWeigh? - it's 0.0006

\subsubsection{Combined Approaches}

While the approaches discussed so far each build on the individual ideas of identifying a clean subset or delaying memorization, many current approaches in fact combine multiple of such ideas in multi-stage pipelines \cite{liang2020bond,yu2021fine, wang2022promix}. As representative of such approaches, we evaluate BOND~\cite{liang2020bond} and meta self-refinement (MSR)~\cite{zhu-etal-2023-meta}, both of which combine pseudo-labeling in a student-teacher setup and confidence-based sample selection.

%\textbf{BOND.} It is a two-stage framework that follows a regular transformer fine-tuning objective in the first stage. The second stage consists of a teacher-student setup: the student model is first initialized by the model learned in the first stage and trained using soft or hard pseudo-labels, produced by the same model (teacher model). Then, the teacher model is updated from the student model in the previous iteration to generate a new set of pseudo-labels for the next iteration of the student model. In addition, samples are selected based on the prediction confidence of the student model.% to further improve the quality of pseudo-labels. 
\todo[]{moved some implementation details to appendix}

\noindent
\textbf{No upper bound for pseudo-labeling.} We cannot derive a separate upper bound for pseudo-labeling, as the best case scenario here would mean that all noisy labels are replaced by correct labels, which is the same as training on the \noiseclean~dataset.%. This means that the upper bound for pseudo-labeling is the same as oracle subset.

\subsubsection{Additional Clean Data}

We include a further upper bound for the scenario in which a small amount of high quality noise-free data is available. This is inspired by the extensive analysis of the use of clean validation data in \citet{zhu-etal-2023-weaker}. Here, after first training on the noisy training set, they use a small clean dataset to continue fine-tuning the model. We include this upper bound to measure the accuracy gains that may be achieved if one were to invest effort in manually annotating additional noise-free data. 

%In the previous sections, we derive the oracle subset and oracle stopping upper bounds. Additionally, inspired by the extensive analysis of the use of clean validation data in \cite{zhu-etal-2023-weaker}, we provide an upper bound for when additional clean data is available. This setting assumes a small set of high quality examples. After an initial stage of training on the noisy training set, the small clean dataset is used to continue fine-tuning the model. 

%\vspace{1mm}

\subsection{Results} Table \ref{exp3_conll_results} summarizes the evaluation results. We make the following observations:

\noindent
\textbf{Identifying a clean subset has highest potential.} The upper bound of training only the clean subset of each noisy split (see "Oracle subset" in  Table~\ref{exp3_conll_results}) achieves the best scores of all upper bounds. This makes intuitive sense as training is performed only over fully clean sentences, albeit a smaller subset of the full training data as all noisy sentences are filtered out. Similarly, we find strong improvements for the "Additional clean data" upper bound. Oracle stopping, on the other hand, does not achieve the same level of performance as the oracle subset, only slightly outperforming the FLERT baseline. This is in line with our findings in Experiment 2 that the early-learning generalization phase is skipped when training with real noise. This indicates that noise-robust learning approaches that target early stopping  have little potential.

\noindent
\textbf{Small benefit of noise-robust approaches.} Evidently, there is no single best approach for all noise types. For each noise type, at least one noise-robust approach outperforms the baseline, however on average most of them are comparable to it. Only MSR outperforms the baseline averaged over all noise types, bringing improvements for \noisecrowdbest, \noisedistant, \noiseweak~and \noisellm. Additionally, L2R works well for \noisellm~noise and BOND for \noisecrowd. Still, the performance is far below the upper bound. This raises the issue of trade-offs of existing noise-robust learning approaches, since they often require additional hyperparameter tuning or incur computational costs, but only lead to slight improvements in the presence of real noise. % not sure if this is the right conclusion

\section{Ablation: \benchmark{} for German}
\label{sec:ablation_german}

Using the German sentences in CoNLL-03, we created a noisy label benchmark for German to confirm our findings for a different language. Following the English counterpart described in Section \ref{benchmark}, it consists of (1) a noise-free test split to evaluate trained models and (2) three variants of the training split, where two are annotated with different types of noise and one is the ground truth. The two types of noise include \noiseexpert~labels, with 16.2\% noise and \noisellm~labels, with 54\% noise. More details can be found in Appendix \ref{appendix-german-benchmark-overview}.

\subsection
{Experimental Results}

\input{tables/conll_german_results}

Table \ref{exp3_conll_results_german} shows the results of the noise-robust approaches and upper bounds when training on the German datasets. More experimental details and results can be found in Appendix \ref{appendix:german_experiments}.

\noindent
\textbf{Oracle subset score reaches an upper limit.} Regarding the upper bounds, we see that the performance of the oracle subset of \noiseexpert~and \noisellm~is close, meaning that the 4000\footnote{See Appendix \ref{appendix:oracle_subset_size} for the size of the oracle subset.}  
clean sentences in the \noisellm~subset are already enough to reach an F1 score over 82. Despite having more samples, the \noiseexpert~subset does not result in a much higher score. This could signify that the remaining sentences, not included in the \noiseexpert~subset, are difficult examples necessary to properly learn the task. 

\noindent
\textbf{Poor performance of noise-robust approaches.} Regarding the noise-robust approaches, only Confident Learning is able to match and slightly outperform the baseline. All other methods mainly perform poorly on the German dataset, even below the baseline, with the exception of the improvement brought by MSR on the \noisellm~dataset.

\section{Related Work}
\vspace{1mm}

% \textbf{Training with partially incorrect labels.} Some works focus on WS, some on DS, some on simulated noise. Their pipelines share similar characteristics. 

% \textit{Label noise robustness for NER.}
% \todo[inline]{first paragraph could be removed if neccessary, we're already mentioning most of this stuff}
% \textbf{Learning with noisy labels.} There have been various studies proposing approaches to mitigate the negative effects of partially incorrect annotations and enforce better generalization. These studies usually concern with a single source of noise, such as crowdsourced annotations~\cite{wei2021learning,klie2023annotation}, distant~\cite{hedderich2021analysing} or weak supervision approaches~\cite{zhang2021wrench}, however they share similar ideas, such as sample selection~\cite{northcutt2021confident, liu-etal-2021-noisy-labeled}, sample reweighing \cite{Wang2019CrossWeighTN, yu2021fine} %bond
% and co-training~\cite{zhou2021learning, liang2020bond}. % - we argue that an evaluation across noise types and levels would be beneficial.

There are a few benchmarks for learning with label noise and related areas. The WRENCH benchmark \cite{zhang2021wrench} focuses only on weak supervision labels for multiple tasks, with the emphasis on combining multiple weak labelling sources. \citet{klie2023annotation} compare a large number of methods for the detection of annotation errors. Multiple tasks are included, including NER on CoNLL-03, where they evaluate the detection of expert errors, concluding that most approaches are not successful at this. Similarly, \citet{chong2022detecting} evaluate annotation error detection on datasets with noise only from crowdsourced labels, for part-of-speech tagging and natural language inference tasks. \citet{liu-etal-2022-noise} propose a benchmark for text classification under label noise, where they re-annotate an existing sentiment classification dataset and construct noisy label sets according to annotator disagreements; however, they do not publish these label sets. NoisyWikiHow, a benchmark for intention identification has also been presented \cite{wu-etal-2023-noisywikihow}, where the authors propose a method to simulate realistic noise that imitates human errors by producing heterogeneous and instance-dependent errors. For NER in Estonian, \citet{hedderich2021analysing} introduce the NoisyNER, which includes multiple noise levels obtained from distant supervision approaches with varying quality. MultiCoNERv2 \cite{fetahu-etal-2023-multiconer} addresses textual noise in the input data itself (e.g. typos), instead of label noise.

% next iteration: \todonotes[]{extend rel work with other studies that look at real and simulated noise (for NER and not), look at memorization }
% \textbf{Simulated noise.} List papers with simulate NER noise and how they do it. 

% \textbf{Memorization of label noise in LLMs./ Real vs simulated noise. } In \cite{tanzer-etal-2022-memorisation} they look at memorization of simulated noise. In \citet{chong2022detecting} and \cite{Zhu2022IsBR}- they don't actually look at training curves - but they investigate something similar as us, but for (a bunch of) other tasks (not NER) and they also show that models are much less robust to real noise than a bunch of simulation models. In \citet{chong2022detecting} real noise is crowdsourced labels, in \cite{Zhu2022IsBR} real noise is Distant. 

% (Tanzer 2022) However, they conducted their experiments with simulated uniform noise. Our goal is to analyze this behavior in a more realistic scenario and to find out if the real label noise affects this memorization dynamics differently.

% Other papers do look at memorization in real vs simulated noise and make the same conclusion as us in Exp 2, but in image classification datasets (cite Clothing1M). 

\section{Conclusion}
\vspace{1mm}

In this paper, we address the issue of label noise in the NER task. We introduce a new benchmark, based on the commonly used NER dataset CoNLL-03, for evaluating the impact of 6 distinct types of real label noise on the same set of sentences, with varying degrees of difficulty. 

We demonstrated that real noise causes transformer-based language models to immediately memorize the noise pattern, making real label noise a more challenging problem than simulated label noise, even in the case of oracle class-dependent noise informed by the characteristics of real noise. 

We further presented an evaluation of popular noise-robust learning approaches. % Our experiments indicate that the correctly-labeled subset in noisy-labeled data has higher potential than what current methods can achieve, leaving room for improvement. 
Our experiments indicate that current methods fall far short of what can  potentially be achieved on the noise types in \benchmark{} and that approaches that focus on automatically identifying a clean subset of labels have the highest potential.  We hope that \benchmark{} aids other researchers in the further development of more effective noise-robust approaches. 

\section*{Limitations}

This paper focuses on the scenario when the entire available dataset could be noisy and we do not have access to a small, high-quality labelled, data subset. While this is a certainly scenario which reflects a large number of real-world cases, it could be argued that in some situations it is realistic to have the resources to ensure a subset of the data is clean, with high-quality annotations. However, when this is the case, \citet{zhu-etal-2023-weaker} showcased that this clean data would be better utilized by directly fine-tuning the models on it, instead of using it for validation. Therefore, we argue that this alternative setup is not particularly useful for the evaluation of label-noise-robust approaches. 

% \section{Ethical considerations}
% (copied from \cite{Zhu2022IsBR}, something similar)
% Noisy labels are a cheaper source of supervision.
% This could make it easier to use machine learning
% for improper use cases. However, it also opens
% up NLP methods for low-resource settings such as
% under-resourced languages or applications devel-
% oped by individuals or small organizations. It can,
% therefore, be a step towards the democratization of
% AI.

\section*{Acknowledgements}

We thank all reviewers for their valuable comments. Elena Merdjanovska and Alan Akbik are supported by the Deutsche Forschungsgemeinschaft (DFG, German Research Foundation) under Germany’s Excellence Strategy – EXC 2002/1 “Science of Intelligence” – project number 390523135. Ansar Aynetdinov and Alan Akbik are supported by the Deutsche Forschungsgemeinschaft (DFG, German Research Foundation) under Emmy Noether grant “Eidetic Representations of Natural Language” (project number 448414230).

% Place all acknowledgments (including those concerning research grants and funding) in a separate section at the end of the paper.

\bibliography{noisebench}

\appendix

\input{additional_content/appendix_german_benchmark_camera_ready}

\input{additional_content/appendix_implementation_details}

\input{additional_content/appendix_clean_subsets_size}

\input{additional_content/appendix_uniform_noise}

\input{additional_content/appendix_memorization_additional_plots}

\input{additional_content/appendix_more_experiments_memorization}

\input{additional_content/appendix_analysis_of_predictions_baseline_camera_ready}

\end{document}

%% file: figures/text_example_figure.tex
\begin{figure*}
    \centering
    \vspace{-1mm}
    \begin{subfigure}[t]{0.264\textwidth}
        \centering
    	\includegraphics[width=\linewidth]{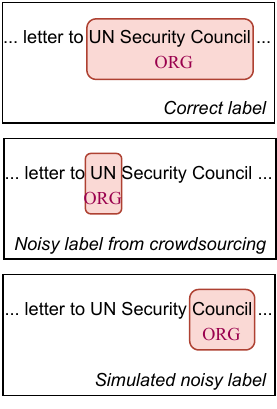}
    	\caption{Example of a \textbf{partial match} error induced by noise. Real noise makes a plausible mistake by labeling "UN" as ORG (organization), whereas simulated noise implausibly caused "Council" to be labeled.}
    	\label{fig:figure_partial_matches}
    \end{subfigure} \hfill
    \begin{subfigure}[t]{0.37\textwidth}
        \centering
    	\includegraphics[width=\linewidth]{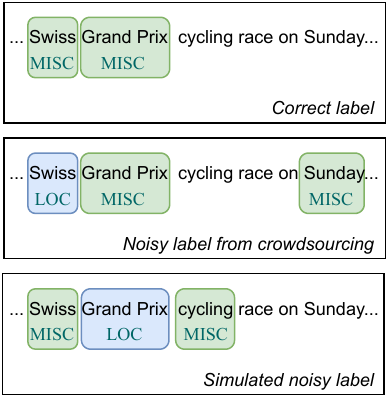}
    	\caption{Examples of \textbf{type} and \textbf{non-entity} errors induced by noise. Real noise makes a plausible mistake by labeling "Swiss" as a LOC (location), whereas simulated noise implausibly labels "Grand Prix" as LOC. Real noise makes a plausible non-entity mistake by labeling "Sunday", whereas simulated noise labels "cycling".
     %Similarly, : The real noisy label shows a more plausible \textbf{type} error, with (\textit{Swiss} labeled as LOC - location) and \textbf{non-entity} error (\textit{Sunday} labelled as a MISC - miscellaneous entity) than the simulated ones. 
     }
    	\label{fig:figure_missing}
     \end{subfigure} \hfill 
     \begin{subfigure}[t]{0.286\textwidth}
        \centering
		\includegraphics[width=\linewidth]{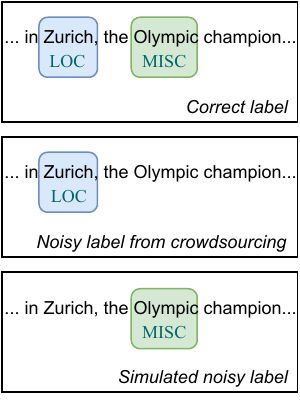}
    	\caption{Example of a \textbf{missing} mention induced by noise. Real noise causes a plausible omission ("Olympic"), whereas simulated noise omits a trivial entity annotation ("Zurich")}
    	\label{fig:figure_misc}
     \end{subfigure}
     \vfill
  
\caption{Examples of text snippets with correct labels (top row) and two types of noise: Real noise from crowdsourcing (middle row) and simulated class-dependent noise (bottom row). This introduces different types of errors: (a) partial matches of correct entity mentions, (b) a wrong type and a non-entity mention and (c) a missing entity. We qualitatively find real noise to be more plausible than simulated noise.}

\vspace{-1.5mm}
\label{fig:example_sentence_figure}
\end{figure*}

%% file: tables/noisebench_overview_shares.tex
\begin{table*}[]
\vspace{-1mm}
\centering
\small
\setlength{\tabcolsep}{4pt}
\begin{tabular}{@{}lrrrrrrrrr@{}}
\toprule
 & &  & & \multicolumn{2}{c}{\textit{\#Entities}} &\multicolumn{4}{c}{\textit{\%Errors}} \\
  \cmidrule(lr){5-6}   \cmidrule(lr){7-10}  
 \textit{Noisy train split} & \textit{\%Noise} & \textit{F1}\scriptsize{token} & \textit{F1\scriptsize{entity}} & \textit{Total} & \textit{Correct} & \textit{Missing (FN)} & \textit{Non-entity (FP)} & \textit{Type} & \textit{Partial} \\ 
\midrule %\multicolumn{3}{@{}l}{\textit{Noisy training splits}}\ &  &  &  &  &  &  &  &  &  \\
% \noiseclean & 0 & 100 & 100 & 100 & 100 & 9,685 & 9,685 & 0 & 0 & 0 & 0 \\
\noiseexpert & 5.5 & 99.0 & 94.5 & 9,644 & 9,129 & 10.0 & 2.8 & 74.0 & 13.3 \\
\noisecrowdbest & 15.3 & 96.7 & 84.7 & 8,607 & 7,751 & 59.6 & 8.7 & 17.0 & 14.7 \\
\noisecrowd & 36.6 & 92.3 & 63.4 & 7,188 & 5,352 & 61.9 & 10.2 & 16.0 & 11.9 \\
\noisedistant & 31.3 & 92.9 & 68.7 & 7,329 & 5,846 & 65.4 & 10.5 & 12.9 & 11.1 \\
\noiseweak & 40.4 & 91.9 & 59.6 & 10,640 & 6,058 & 17.4 & 34.6 & 36.3 & 11.8 \\
\noisellm &  45.6 & 87.4 & 54.4 & 11,349 & 5,726 & 22.5 & 45.4 & 28.3 & 3.7 \\
% \midrule 
% \textit{Test split} &  &  &  &  &  &  &  &  &  &  &  \\
% \noiseclean~Test & 0 & 100 & 100 & 100 & 100 & 5,725 & 5,725 & 0 & 0 & 0 & 0 \\
\bottomrule
\end{tabular}
\caption{\label{conll_noise_shares_table} Overview of the noisy training splits in \benchmark. The table shows the noise level, the micro-averaged token-level F1 score (\textit{F1}\begin{scriptsize}token\end{scriptsize}), micro-averaged entity-level F1 (\textit{F1}\begin{scriptsize}entity\end{scriptsize}), the number of entities (\textit{Total}), number of correct entities (\textit{Correct}) and share of each error type: missing mentions (\textit{Missing (FN)}), non-entity mentions (\textit{Non-entity (FP)}), wrong type (\textit{Type}) and partial matches (\textit{Partial}). All metrics are in reference to the \noiseclean~split.} 
\vspace{-2mm}
\end{table*}

%% file: tables/simulated_vs_real_noise_shortened.tex
\begin{table}[!b]
\setlength{\tabcolsep}{4pt}
\centering
\small
\begin{tabular}{lccccc}
\toprule
%  &  & \multicolumn{2}{c}{ F1 }\\ \midrule
% \noiseclean &  & \multicolumn{2}{c}{93.99 ± 0.04}\\ \midrule
  &  \multicolumn{2}{c}{ Real noise } & \multicolumn{2}{c}{ Simulated noise } & $\Delta$ \\ \cmidrule(lr){2-3} \cmidrule(lr){4-5} \cmidrule(lr){6-6} 
 & \textit{\%Noise} & \textit{F1} & \textit{\%Noise} & \textit{F1} & \textit{F1} \\ \midrule
 \noiseclean & 0 & 94.0 \scriptsize{±0.0} & - & - & - \\
\noiseexpert & 5.5 & 89.8\scriptsize{±0.2} & 5.9 & 93.7\scriptsize{±0.2} & 3.9 \\
\noisecrowdbest & 15.3 & 86.7\scriptsize{±0.3} & 17.9 & 88.9\scriptsize{±0.4} & 2.2 \\
\noisecrowd & 36.6 & 70.5\scriptsize{±0.6} & 41.3 & 72.4\scriptsize{±1.0} & 1.8\\
\noisedistant & 31.3 & 70.8\scriptsize{±0.1} & 39.2 & 74.5\scriptsize{±0.4} & 3.7 \\
\noiseweak & 40.4 & 65.9\scriptsize{±0.4} & 41.2 & 63.1\scriptsize{±0.8} & -2.8 \\
\noisellm & 45.6 & 62.6\scriptsize{±0.4} & 47.2 & 68.6\scriptsize{±1.3} &6.0 \\ 
Average &  & 74.4\scriptsize{±0.3} & & 76.9\scriptsize{±0.7} & 2.5 \\ \bottomrule
\end{tabular}
\caption{\label{conll_baselines_real_simulated} F1 scores on the \noiseclean~Test split of the baseline FLERT approach, fine-tuned on different noisy variants of the training set. The scores are averages of 3 runs. The column  $\Delta$ (difference) refers to the difference in F1 score on the test split when training on a dataset with real noise compared to simulated class-dependent noise.
} %The scores are averaged over 3 runs. Training on the clean training set has a score of 93.99}%\scriptsize{±0.04}}
\vspace{-2mm}
\end{table}

%% file: figures/seen_vs_unseen_plots.tex
\begin{figure}[!b]
\centering
\begin{subfigure}[]{0.41\textwidth}
\centering
 \includegraphics[width=\linewidth]{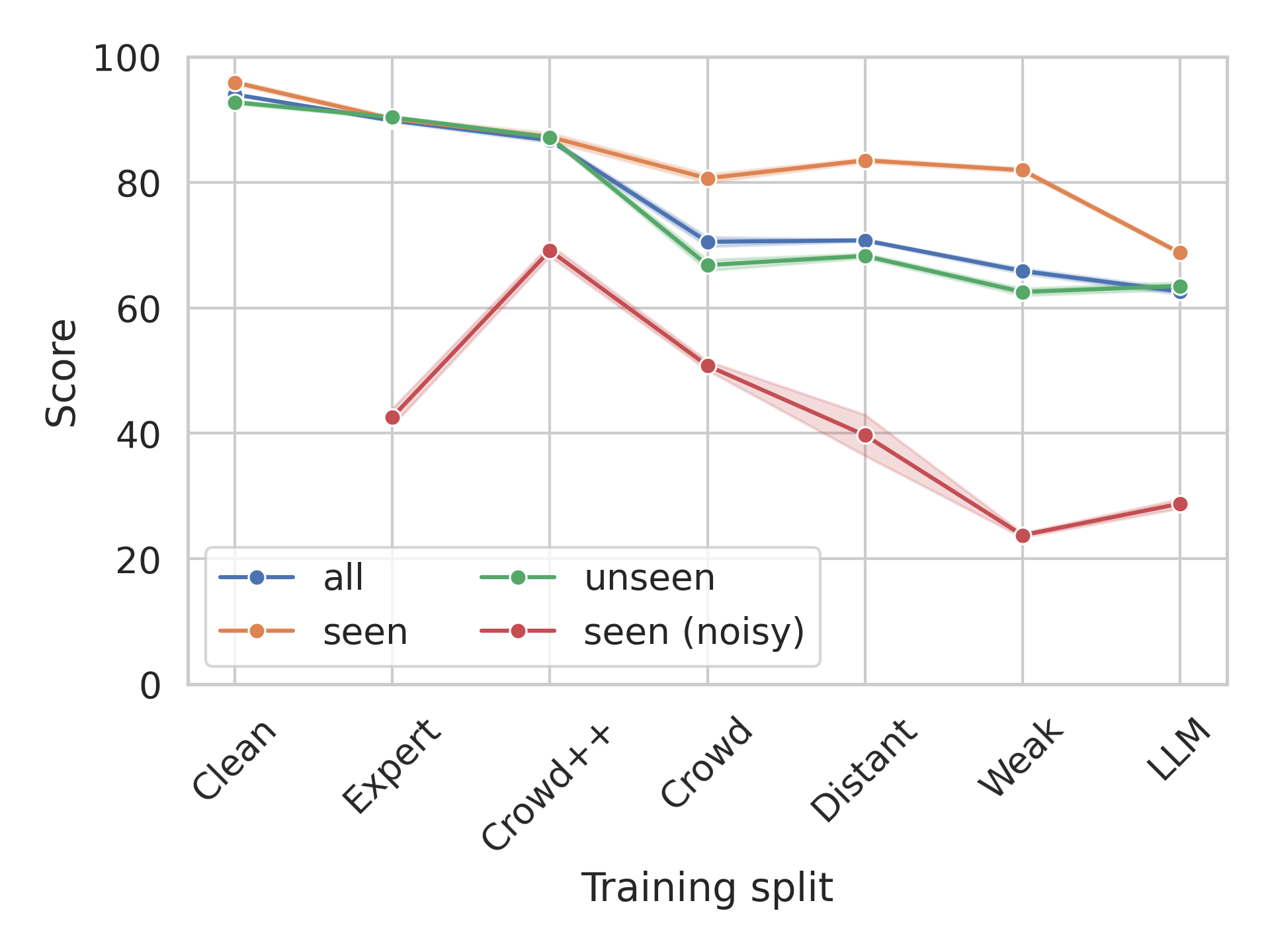} 
\caption{Real noise}
\label{fig:seen_vs_unseen_real}
\end{subfigure} 
\begin{subfigure}[]{0.41\textwidth}
\centering
 \includegraphics[width=\linewidth]{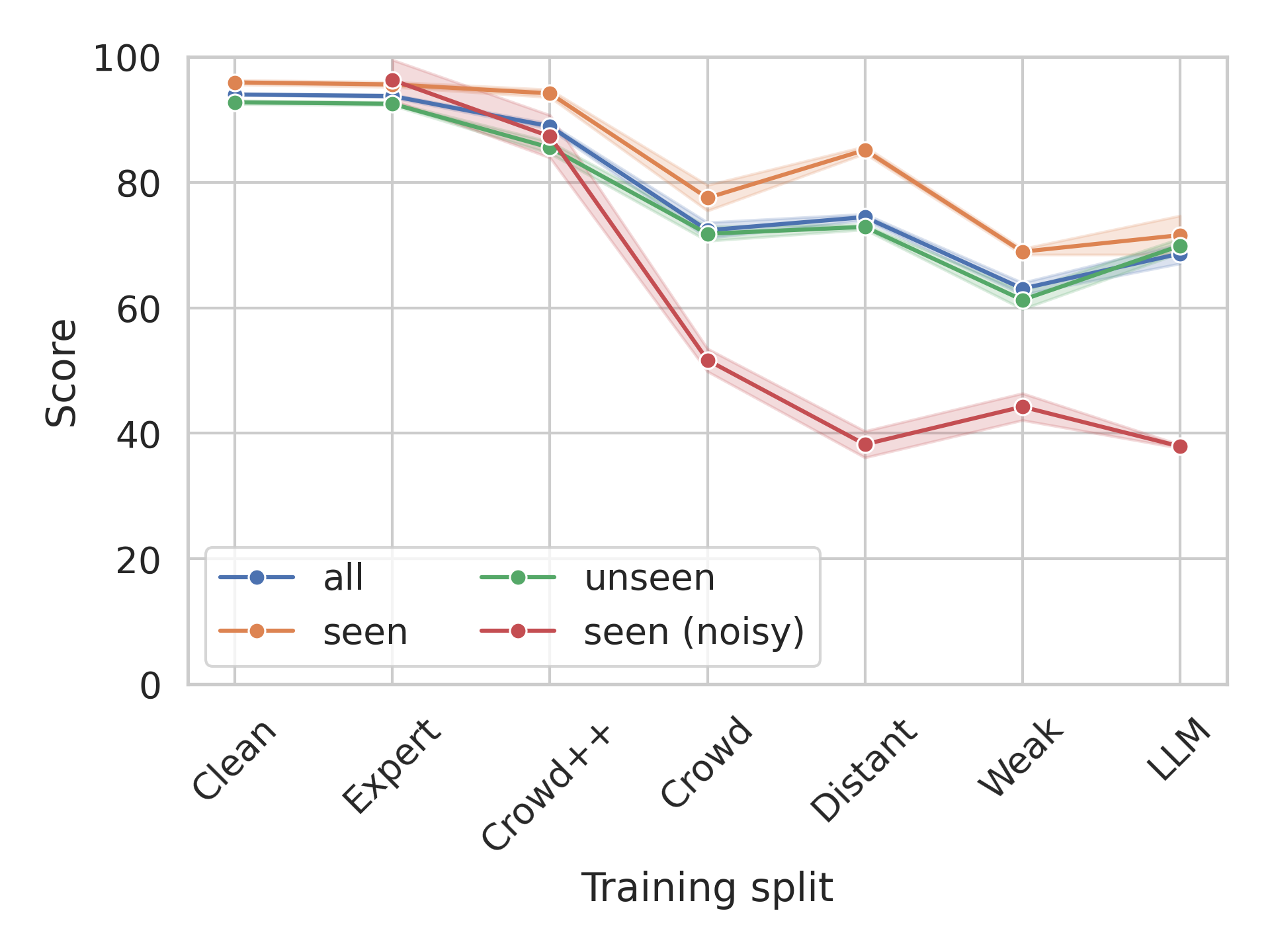} 
\caption{Simulated class-dependent noise}
\label{fig:seen_vs_unseen_simulated}
\end{subfigure} 
\caption{F1 scores on different subsets of entities in the test set: all, seen (clean), seen (noisy) and unseen.}
\label{fig:seen_vs_unseen}
\vspace{-2mm}
\end{figure} 

%% file: figures/memorization_plots_grid.tex
\renewcommand{\tablename}{Figure}
\setcounter{table}{2} 
\begin{table*}
\centering
\begin{tabular}{m{0.1cm} >{\centering\arraybackslash}m{4.6cm} >{\centering\arraybackslash} m{4.6cm} >{\centering\arraybackslash}m{4.6cm}}
 & \hspace{1cm}\noiseexpert & \hspace{1cm}\noisecrowdbest & \hspace{1cm}\noisedistant \\ 
\rotatebox[origin=c]{90}{Real}\vspace{1.3cm}
& \includegraphics[width=0.31\textwidth]{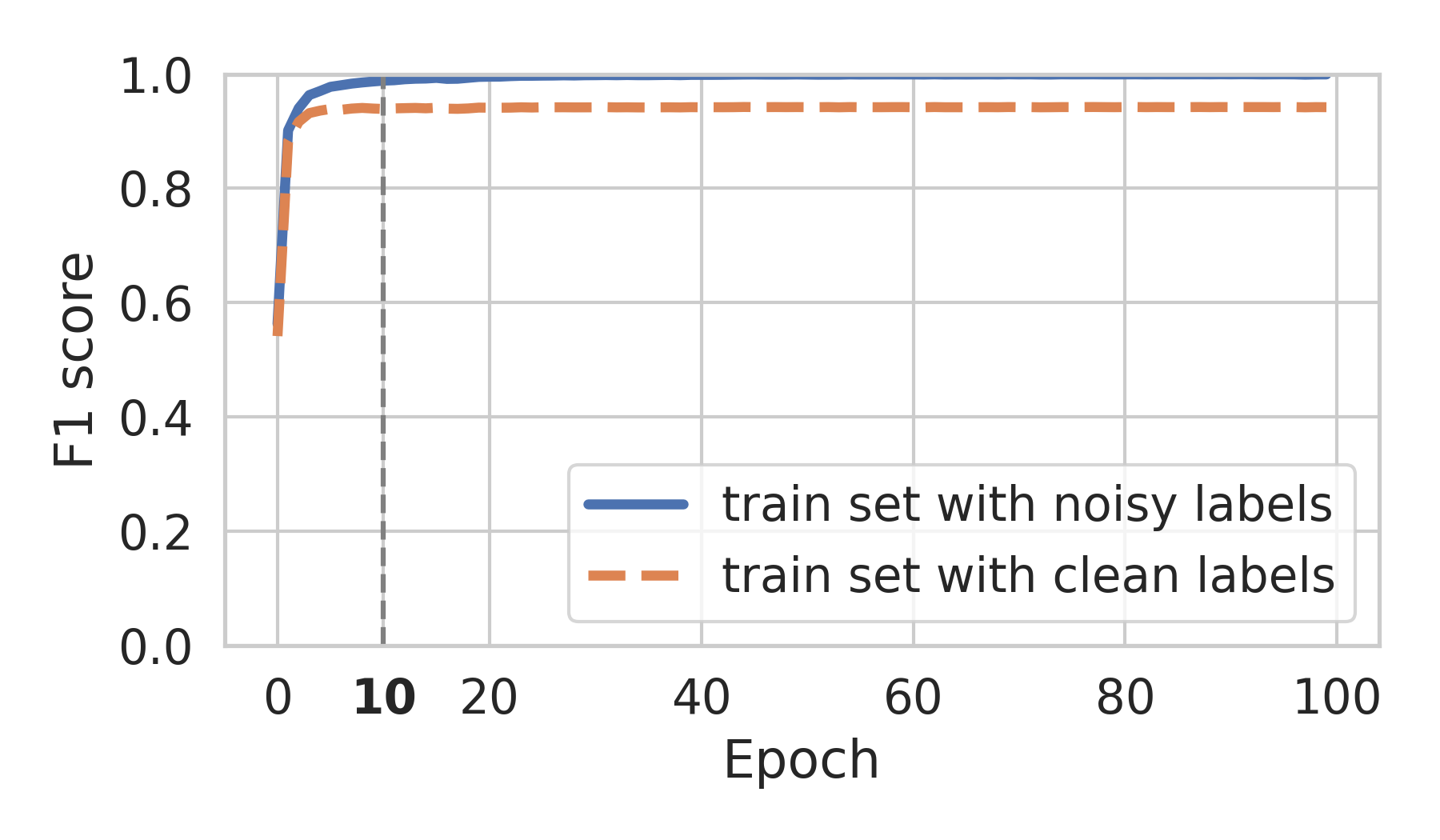} 
\subcaption{Real \noiseexpert~ - 5.5\% noise}
\label{fig:real_original}
& \includegraphics[width=0.31\textwidth]{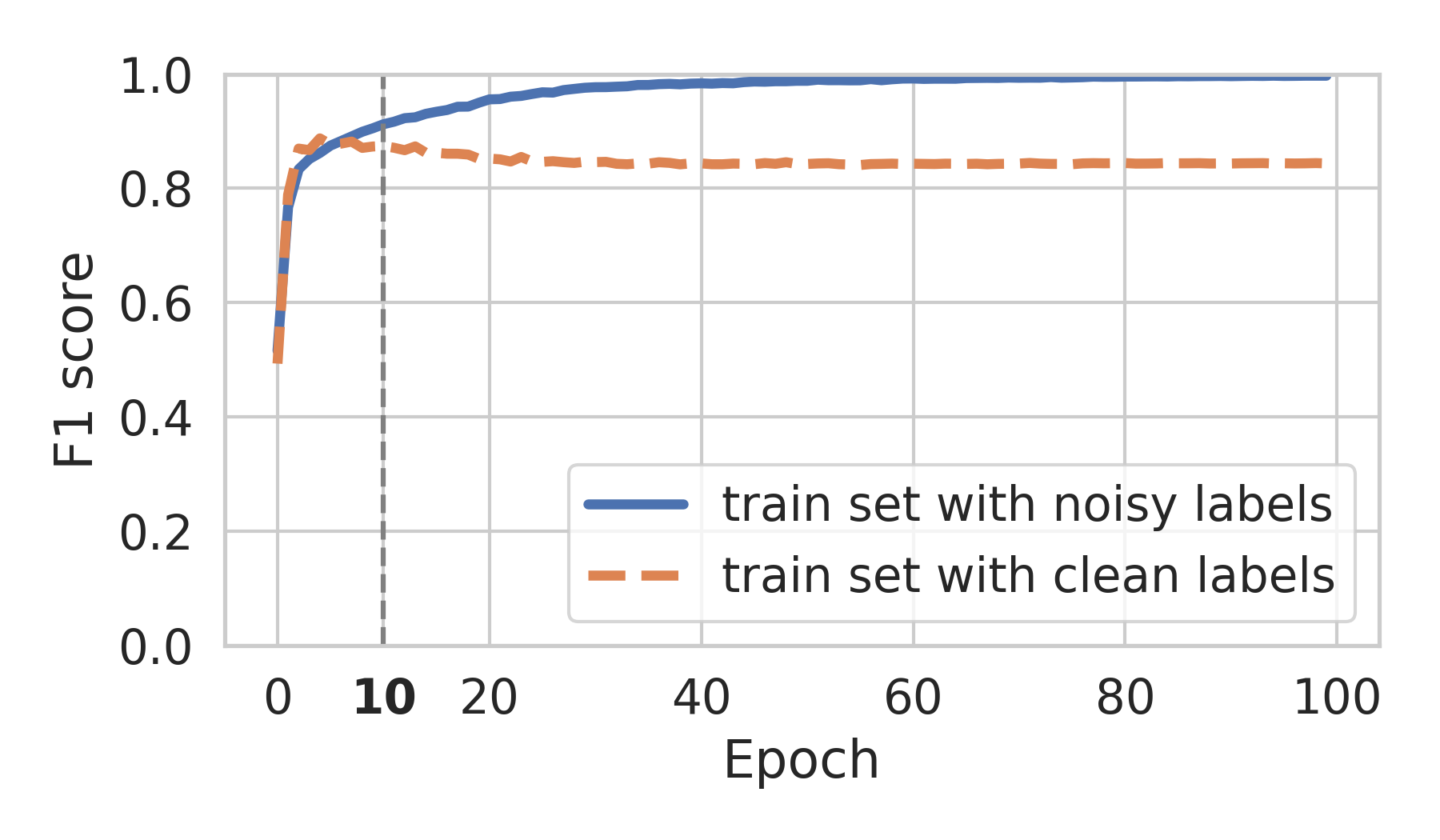} 
\subcaption{Real \noisecrowdbest~ - 18\% noise}
\label{fig:real_mv_oracle}
& \includegraphics[width=0.31\textwidth]{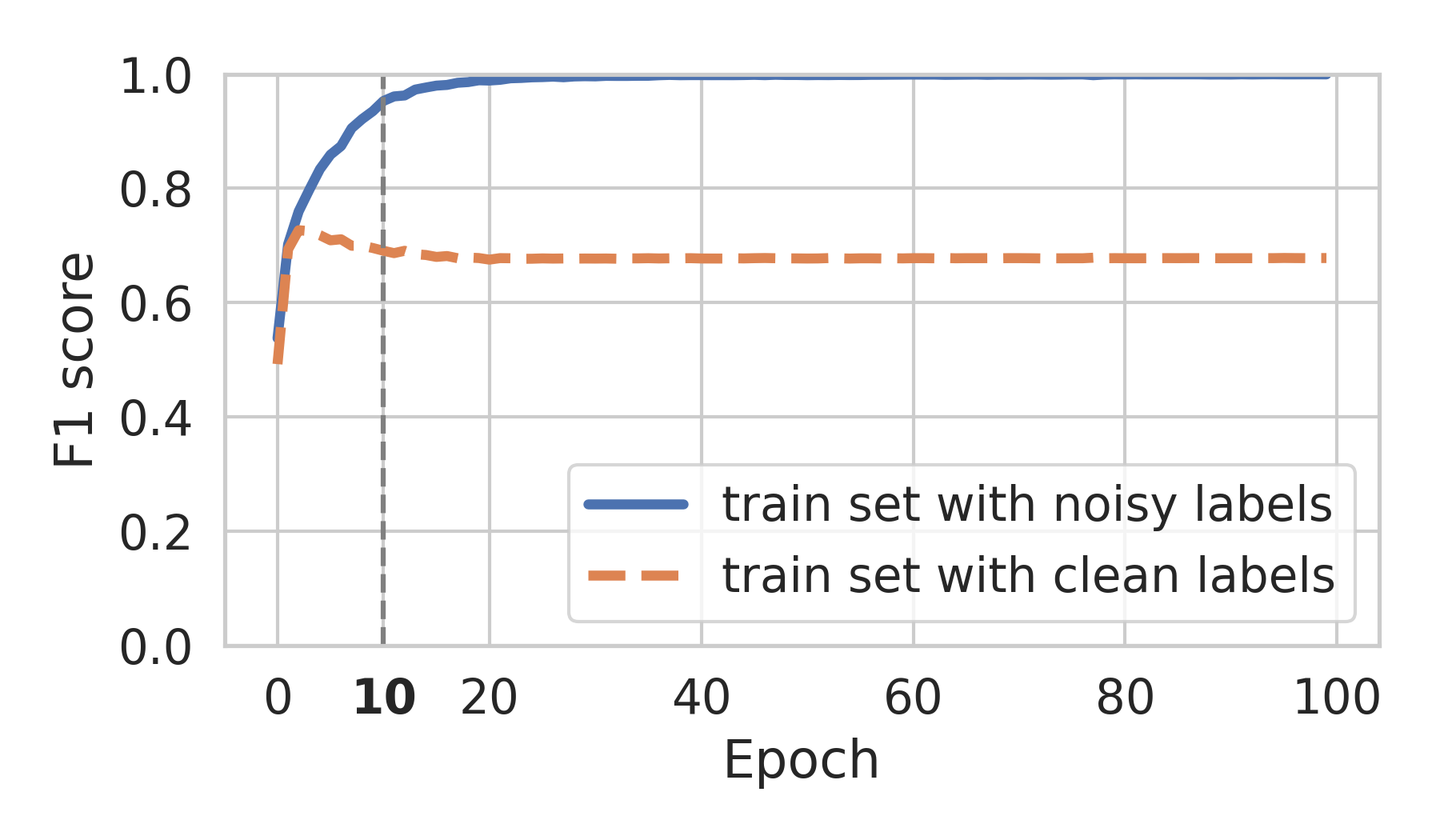}     	
\subcaption{Real \noisedistant~ - 31.3\% noise}
\label{fig:real_bond} \\ 
\rotatebox[origin=c]{90}{Simulated} \vspace{1.3cm}
&  \includegraphics[width=0.31\textwidth]{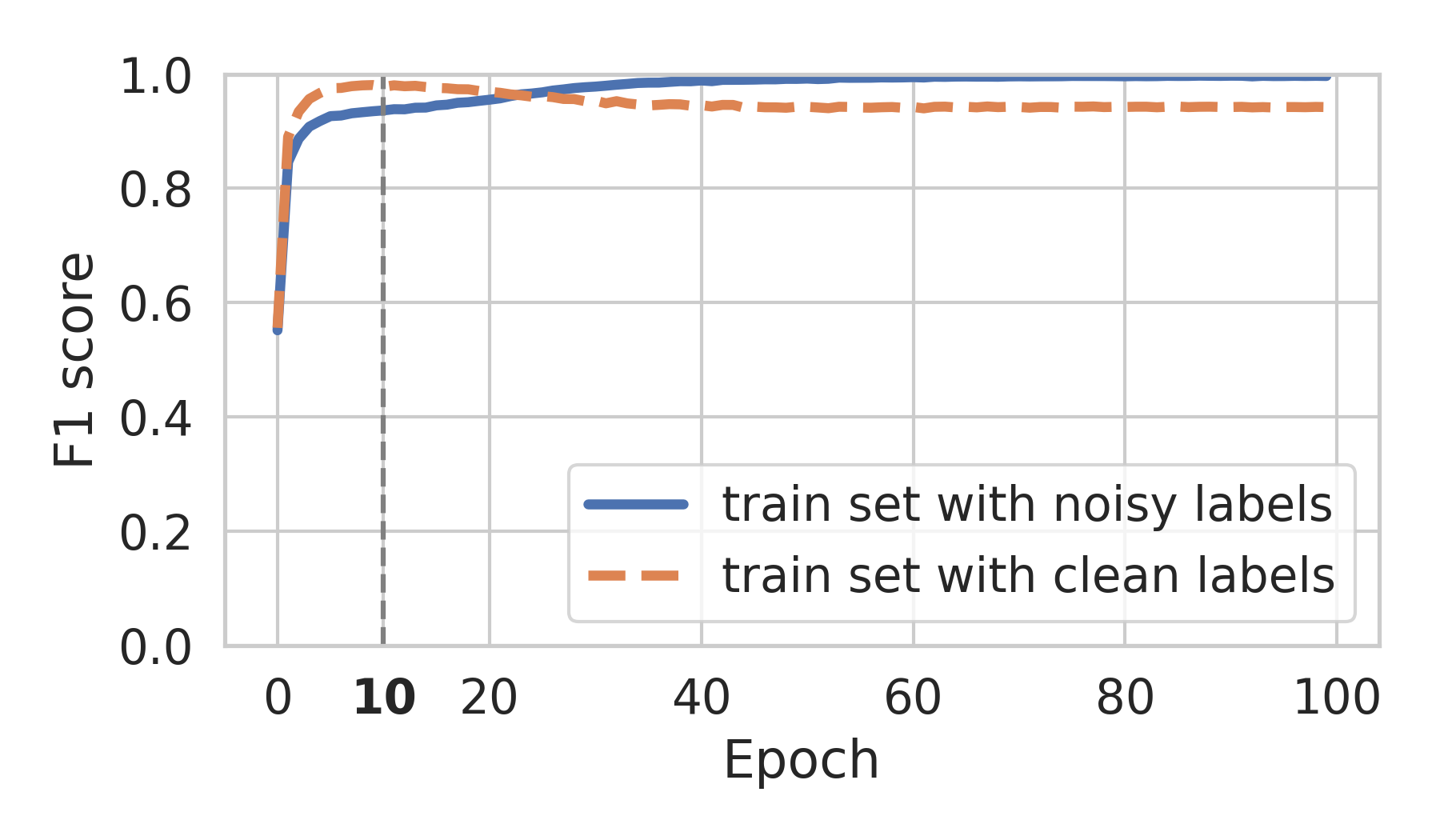} 
\subcaption{Simul. \noiseexpert~ - 6\% noise}
\label{fig:simulated_original}
& \includegraphics[width=0.31\textwidth]{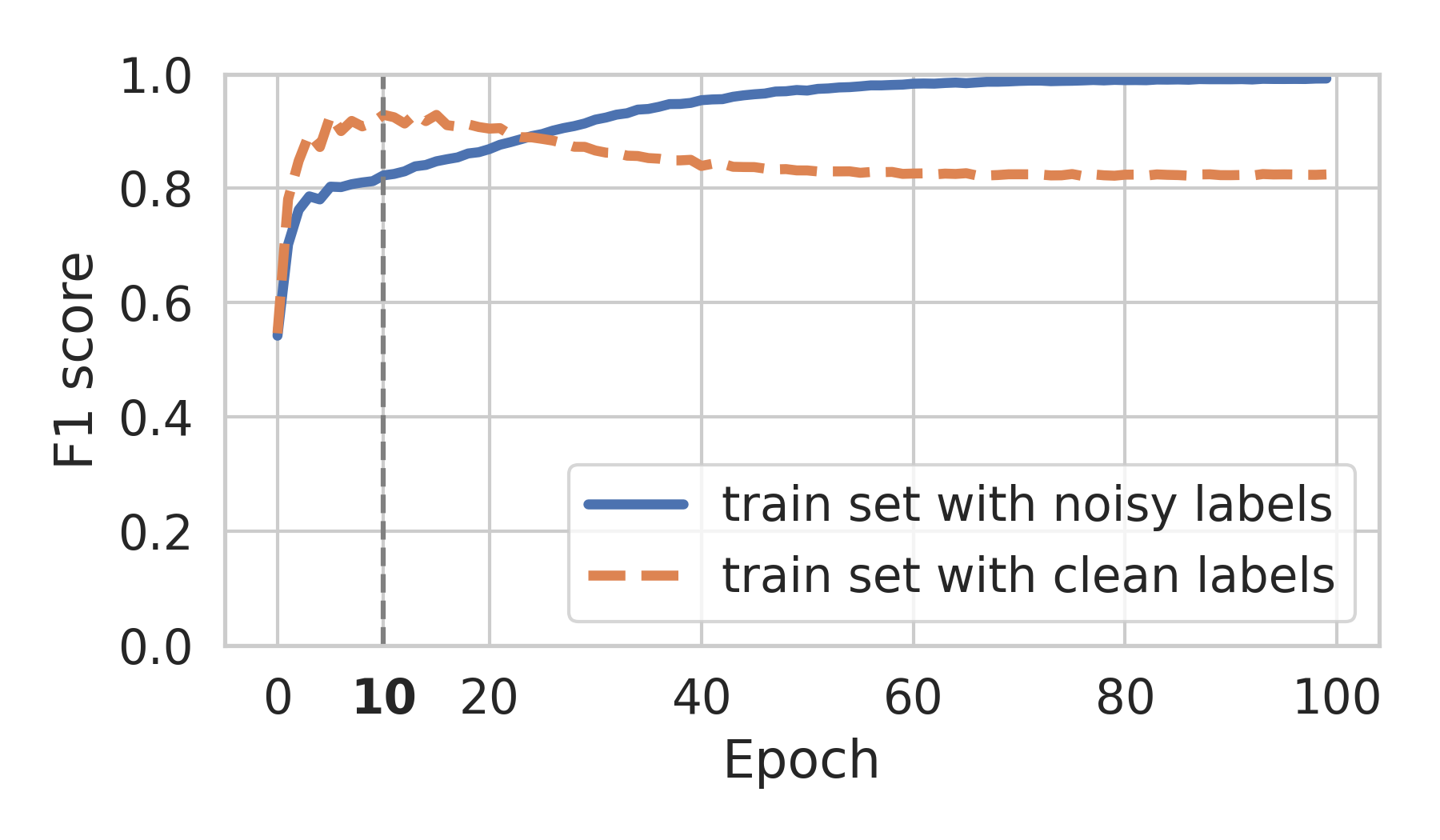} 
\subcaption{Simul. \noisecrowdbest~ - 15\% noise}
\label{fig:simulated_mv_oracle}
& \includegraphics[width=0.31\textwidth]{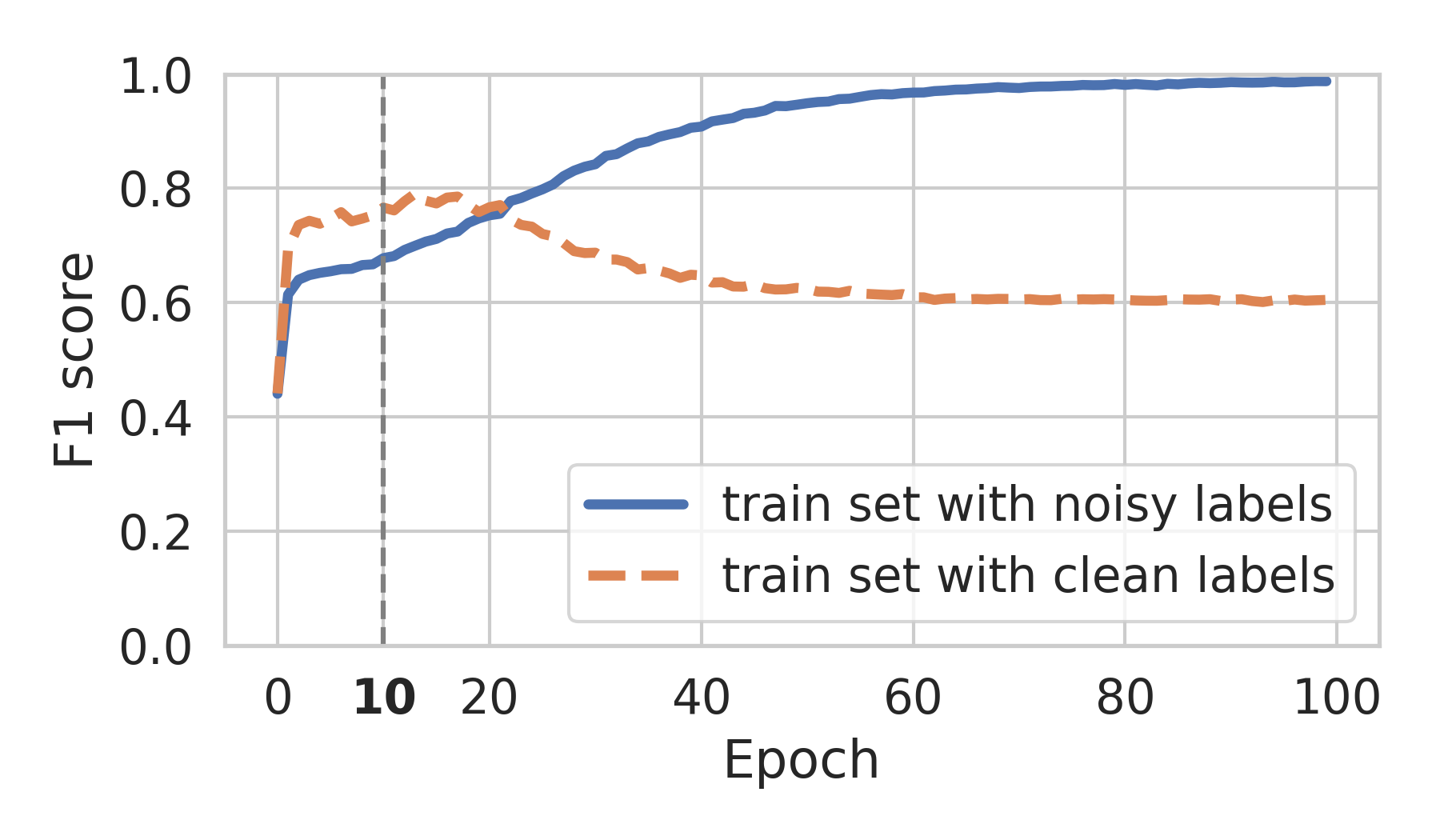} 
\subcaption{Simul. \noisedistant~ - 39\% noise}
\label{fig:simulated_bond}
\\ 
\end{tabular}
\vspace{-4mm}
\caption{\label{fig:memorization_figure}Comparison of model performance during extended training. The top row shows models fine-tuned on label sets with real noise, while the bottom row models fine-tuned on corresponding simulated (class-dependent) noisy labels.  The plots are averages of 3 runs. The graphs for \noisecrowd, \noiseweak~and \noisellm~are shown in Appendix \ref{sec:appendix-memorization}.}
\vspace{-2mm}
\end{table*} 
\renewcommand{\tablename}{Table}
\setcounter{table}{2} 
\setcounter{figure}{3} 

%% file: tables/conll_results.tex
\begin{table*}[]
\small
\centering
\vspace{-2mm}
\setlength{\tabcolsep}{4pt}
\begin{tabular}{p{2.9cm} r r r r r r r r}
%\begin{tabular}{m{0.16\textwidth}>{\centering}m{0.09\textwidth}>{\centering}m{0.09\textwidth}>{\centering}m{0.09\textwidth}>{\centering}m{0.09\textwidth}>{\centering}m{0.09\textwidth}>{\centering}m{0.09\textwidth}>{\centering}m{0.09\textwidth}>{\centering}m{0.09\textwidth}}
\toprule
 & \noiseclean & \noiseexpert & \noisecrowdbest & \noisecrowd & \noisedistant & \noiseweak & \noisellm & Avg. \tabularnewline \midrule
Baseline & 93.99{\scriptsize±0.04} & 89.84{\scriptsize±0.19}& 86.71{\scriptsize±0.29} & 	70.52{\scriptsize±0.62} & 70.75{\scriptsize±0.13} & 	65.87{\scriptsize±0.36} & 	62.60{\scriptsize±0.39} & 77.18 \tabularnewline
% 77.18{\scriptsize±0.29}
 \midrule
 \multicolumn{3}{l}{\textit{Upper bounds}} & & & & & & \tabularnewline
Oracle subset & - &\textbf{90.31}{\scriptsize±0.28}	& \textbf{91.83}{\scriptsize±0.33}	 & \textbf{85.95}{\scriptsize±0.59}	 & \textbf{83.07}{\scriptsize±0.59}	 & \textbf{81.13}{\scriptsize±1.25}	 & \textbf{75.70}{\scriptsize±1.10} & \textbf{85.99}\tabularnewline
% \textbf{85.99}{\scriptsize±0.60}
Oracle stopping & 94.06{\scriptsize±0.07} & 89.88{\scriptsize±0.19} & 87.23{\scriptsize±0.24}	 & 71.04{\scriptsize±0.88}	 & 71.84{\scriptsize±0.61}	 & 66.98{\scriptsize±0.22}	 & 63.64{\scriptsize±0.44} & 77.81 \tabularnewline 
%77.81{\scriptsize±0.38} 
% Oracle weights &  - &  90.75{\scriptsize±0.25}  &  00.00{\scriptsize±0.00} & 00.00{\scriptsize±0.00}  &  00.00{\scriptsize±0.00}  & 00.00{\scriptsize±0.00}  &  00.00{\scriptsize±0.00}  & 00.00 \\

Additional clean data & \textbf{94.14}{\scriptsize±0.18} & 90.04{\scriptsize±0.30} & 89.14{\scriptsize±0.67}	 & 81.70{\scriptsize±0.95}	 & 80.19{\scriptsize±0.73}  & 71.66{\scriptsize±1.66}  & 72.06{\scriptsize±1.32} & 82.70 \tabularnewline 
%77.81{\scriptsize±0.38} 
\midrule
 \multicolumn{3}{l}{\textit{Noise-robust learning}} & & & & & &  \tabularnewline
Confident learning & \textbf{93.71}{\scriptsize±0.31}	& \textbf{90.01}{\scriptsize±0.15}	& 86.53{\scriptsize±0.23} & 69.99{\scriptsize±0.97}	& 71.41{\scriptsize±0.34}	& 65.81{\scriptsize±0.46}	& 61.75{\scriptsize±0.56} & 77.03 \tabularnewline
CrossWeigh & 93.50{\scriptsize±0.12} & 89.68{\scriptsize±0.49}  & 85.01{\scriptsize±0.83} & 64.95{\scriptsize±1.18}  & 70.55{\scriptsize±0.24}  & 66.15{\scriptsize±0.24}  & 60.87{\scriptsize±1.40}  & 75.82
\tabularnewline
L2R & 90.29{\scriptsize±2.12} & 82.10{\scriptsize±4.10}  & 79.91{\scriptsize±2.27} & 67.51{\scriptsize±1.01}  &  65.45{\scriptsize±2.01}  & 63.36{\scriptsize±0.34}  & \textbf{65.29}{\scriptsize±4.15}  & 73.42
\tabularnewline
Co-regularization & 93.65{\scriptsize±0.11}  & 89.55{\scriptsize±0.22} & 86.91{\scriptsize±0.31} & 72.22{\scriptsize±0.73} & 70.45{\scriptsize±0.22} & 65.52{\scriptsize±0.62} & 62.23{\scriptsize±0.76} & 77.22  \tabularnewline

%77.03{\scriptsize±0.43}
BOND & 89.92{\scriptsize±0.71} & 86.78{\scriptsize±0.35} & 86.13{\scriptsize±0.81} & \textbf{74.12}{\scriptsize±0.49} & 73.62{\scriptsize±0.70} & 66.60{\scriptsize±0.36} & 60.99{\scriptsize±0.77}& 76.88 
\tabularnewline
MSR & 92.83{\scriptsize±0.16} & 89.53{\scriptsize±0.48}  & \textbf{88.45}{\scriptsize±1.08} & 68.44{\scriptsize±3.79}  &  \textbf{75.80}{\scriptsize±1.41}  & \textbf{69.48}{\scriptsize±0.32}  & 64.57{\scriptsize±1.22}  & \textbf{78.44}
%CrossWeigh &  &  &  &  & 67.77 &  &  &  
\tabularnewline
\bottomrule
\end{tabular}
\caption{\label{exp3_conll_results} Performance of noise-robust approaches on the \noiseclean~test set, when training on \benchmark{} training split variants. Results are expressed in terms of F1 score. Each score is averaged over 3 runs.}
\vspace{-2mm}
\end{table*}

%% file: tables/conll_german_results.tex
%single column version
\begin{table}[]
\small
\centering
\setlength{\tabcolsep}{2.7pt}
\begin{tabular}{p{2.65cm} r r r r }

\toprule
 & \noiseclean & \noiseexpert & \noisellm & Avg. \tabularnewline \midrule
Baseline & 90.24{\scriptsize±0.2} & 79.02{\scriptsize±0.4}& 57.86{\scriptsize±0.4} & 75.7 \tabularnewline

 \midrule
 \multicolumn{2}{l}{\textit{Upper bounds}} & & & \tabularnewline
Oracle subset & - &\textbf{83.11}{\scriptsize±0.6}	& \textbf{82.72}{\scriptsize±0.6}	 & 82.9 \tabularnewline
Oracle stopping & 90.50{\scriptsize±0.2} & 79.48{\scriptsize±0.3} & 61.81{\scriptsize±0.9} & 77.3  \tabularnewline 
Additional clean data & 89.86{\scriptsize±0.9} & 82.85{\scriptsize±1.7} & 69.50{\scriptsize±1.4} & 80.7 \tabularnewline
\midrule
 \multicolumn{2}{l}{\textit{Noise-robust learning}} & & &  \tabularnewline
Confident learning & 90.00{\scriptsize±0.3}	& \textbf{79.57}{\scriptsize±0.3}& 58.03{\scriptsize±0.2} & 75.9 \tabularnewline
CrossWeigh & \textbf{90.11{\scriptsize±0.3}} & 78.32{\scriptsize±0.3} & 57.50{\scriptsize±0.8} & 75.3 \tabularnewline
L2R & 81.45{\scriptsize±3.1} & 74.14{\scriptsize±0.8} & 53.07{\scriptsize±1.6} & 69.6\tabularnewline
Co-regularization & 88.50{\scriptsize±0.2}  & 78.49{\scriptsize±0.2} & 54.47{\scriptsize±0.5} & 73.8
\tabularnewline

BOND & 86.53{\scriptsize±0.3} & 77.56{\scriptsize±0.5} &  55.89{\scriptsize±0.6} & 73.3  \tabularnewline
MSR & 85.34{\scriptsize±0.5} & 76.42{\scriptsize±0.4} &  \textbf{64.00}{\scriptsize±0.7} & 75.3  \tabularnewline
\bottomrule
\end{tabular}
\caption{\label{exp3_conll_results_german} German variant: Performance of noise-robust approaches on the \noiseclean~test set, when training on each training split variant. Results are expressed in terms of F1 score. Each score is averaged over 3 runs.}
\end{table}

%% file: additional_content/appendix_german_benchmark_camera_ready.tex
\section{\benchmark{} for German: Additional Details}
\label{appendix-german-benchmark}

Following are additional details about the German version of \benchmark{}, as well as results from Experiment 2 for this dataset. 

\subsection{Overview}
\label{appendix-german-benchmark-overview}

The training split contains 12,705 sentences from 553 documents, covering 10,008 entity mentions. The test split contains 3,160 sentences from 155 documents, covering 3,051 entity mentions. 

\subsubsection{Noise-Free Data}
As noise-free labels for the German part of CoNLL-03, we take the updated annotations from 2006\footnote{More details about the revision of the labels can be found in the ner.tz file, downloaded from \url{https://www.clips.uantwerpen.be/conll2003/ner/}, more specifically in the /ner/etc.2006/revision.txt and ner/etc.2006/guide.pdf files.}. This updated label set is considered ground-truth by the research community. 

When compared to the original CoNLL-03 German labels, he most changes in this label set are for the MISC class, most notably the removal of adjectives derived from names. This can alternatively be considered an update in annotation guidelines. With this, it should be noted that for the German dataset we do not have access to labels with verified high quality, as we do for the English counterpart with the CleanCoNLL labels.

\subsubsection{Expert Errors}

Similar as for English, we take the original CoNLL-03 labels \cite{tjong-kim-sang-de-meulder-2003-introduction} as labels with expert errors. This results in a noise share of 16.2\%.

\noindent
\subsubsection{LLM Teacher Models}

Similar as for English, we use GPT3.5 to create a noisy version of the training split annotated by an LLM. This results in a high noise share of 54\%.

\subsubsection{Statistics}

When compared to the noise shares for \noiseexpert~and \noisellm~in \benchmark{} in Table \ref{conll_noise_shares_table} (5.5\% and 45.6\%), the noise shares for German are higher (16.2\% and 54\% respectively). This is due to LLMs performing more poorly on languages other than English, as well as due to fewer research efforts focusing on re-annotating and cleaning the German part of CoNLL-03, resulting in less consistent labels.

We see that most of the errors in German are non-entity mentions. Overall, we also note that type errors and partial matches are much less prominent here (less than 10\%), even though they formed a larger part of the errors in \benchmark. 

\input{tables/noisebench_german_overview_shares}

\subsection{Experiments}
\label{appendix:german_experiments}
\subsubsection{Validation Split}

We take the last 96 documents from the training split to serve as a validation set, corresponding to roughly 17\% of all sentences.

\subsubsection{Experiment 2}
\label{appendix-memorization-german}
We performed Experiment 2 for the German version of \benchmark{}, where the goal is to observe the memorization of label noise. The resulting graphs for both noise types are shown in Figure \ref{fig:memorization_german}. We observe a similar behaviour as in the English part of \benchmark{}, where the real noisy datasets are memorized immediately.

\input{figures/memorization_german}

%% file: tables/noisebench_german_overview_shares.tex
\begin{table*}[]
\centering
\small
\setlength{\tabcolsep}{4pt}
\begin{tabular}{@{}lrrrrrrrrr@{}}
\toprule
 & &  & & \multicolumn{2}{c}{\textit{\#Entities}} &\multicolumn{4}{c}{\textit{\%Errors}} \\
  \cmidrule(lr){5-6}   \cmidrule(lr){7-10}  
 \textit{Noisy train split} & \textit{\%Noise} & \textit{F1}\scriptsize{token} & \textit{F1\scriptsize{entity}} & \textit{Total} & \textit{Correct} & \textit{Missing (FN)} & \textit{Non-entity (FP)} & \textit{Type} & \textit{Partial} \\ 
\midrule %\multicolumn{3}{@{}l}{\textit{Noisy training splits}}\ &  &  &  &  &  &  &  &  &  \\
% \noiseclean & 0 & 100 & 100 & 100 & 100 & 9,685 & 9,685 & 0 & 0 & 0 & 0 \\
\noiseexpert & 16.2 & 98.2 & 83.8 & 11852 & 9156 & 14.5 & 73.0 & 6.1 & 6.4 \\
\noisellm &  54.0 & 92.1 & 46.0 & 16526 & 6102 & 19.8 & 69.9 & 3.2 & 7.0 \\
% \midrule 
% \textit{Test split} &  &  &  &  &  &  &  &  &  &  &  \\
% \noiseclean~Test & 0 & 100 & 100 & 100 & 100 & 5,725 & 5,725 & 0 & 0 & 0 & 0 \\
\bottomrule
\end{tabular}
\caption{\label{conll_noise_shares_table_german} Overview of the noisy training splits in \benchmark~for German. The table shows the noise level, the micro-averaged token-level F1 score (\textit{F1}\begin{scriptsize}token\end{scriptsize}), micro-averaged entity-level F1 (\textit{F1}\begin{scriptsize}entity\end{scriptsize}), the number of entities (\textit{Total}), number of correct entities (\textit{Correct}) and share of each error type: missing mentions (\textit{Missing (FN)}), non-entity mentions (\textit{Non-entity (FP)}), wrong type (\textit{Type}) and partial matches (\textit{Partial}). All metrics are in comparison to the \noiseclean~split.} 
\vspace{-2mm}
\end{table*}

%% file: figures/memorization_german.tex
\begin{figure}[!b]
\centering
\begin{subfigure}[]{0.4\textwidth}
\centering
 \includegraphics[width=\linewidth]{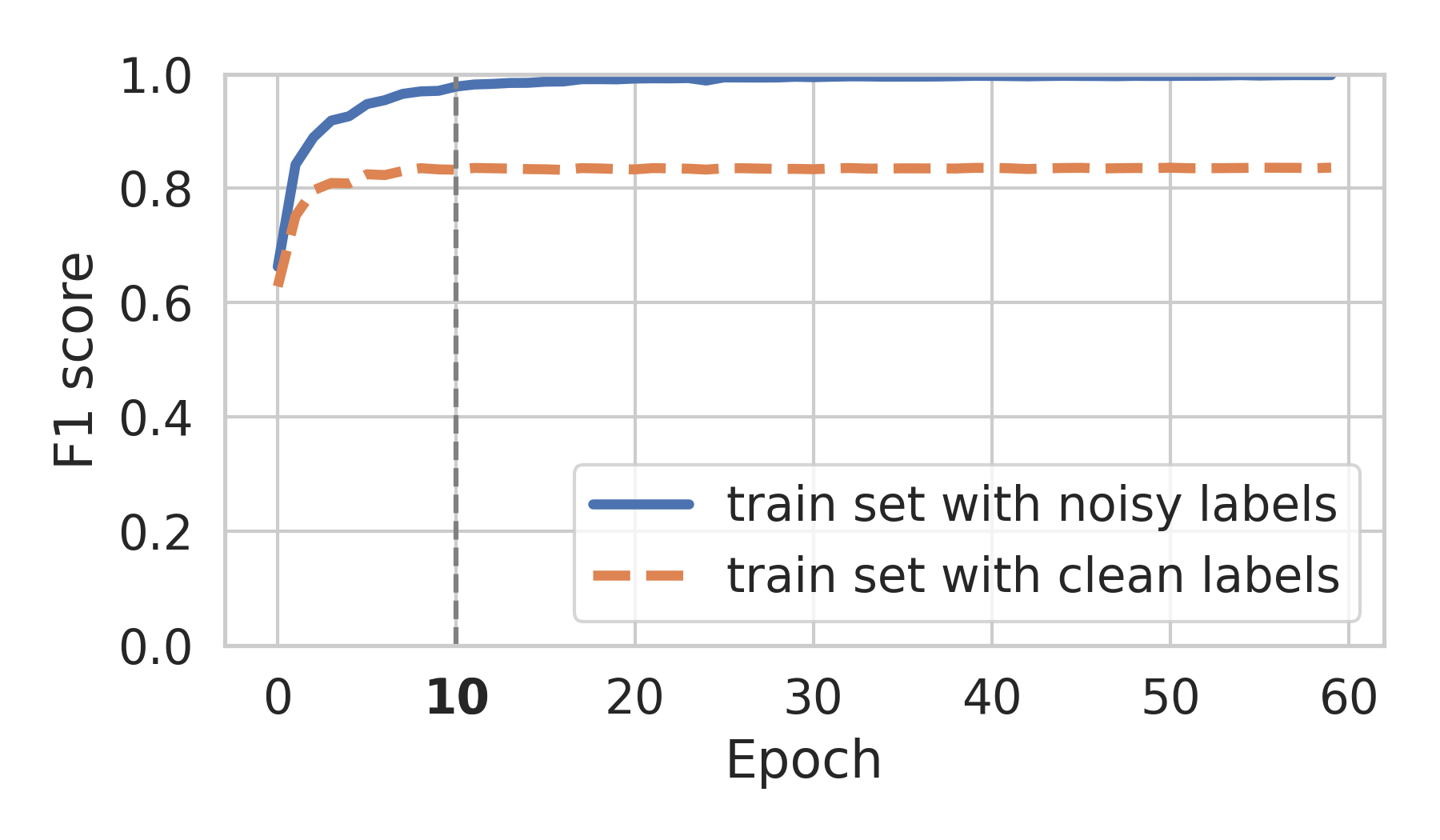} 
\caption{\noiseexpert~- 16.4\% noise}
\label{fig:memorization_german_expert}
\end{subfigure} 
\begin{subfigure}[!htb]{0.4\textwidth}
\centering
 \includegraphics[width=\linewidth]{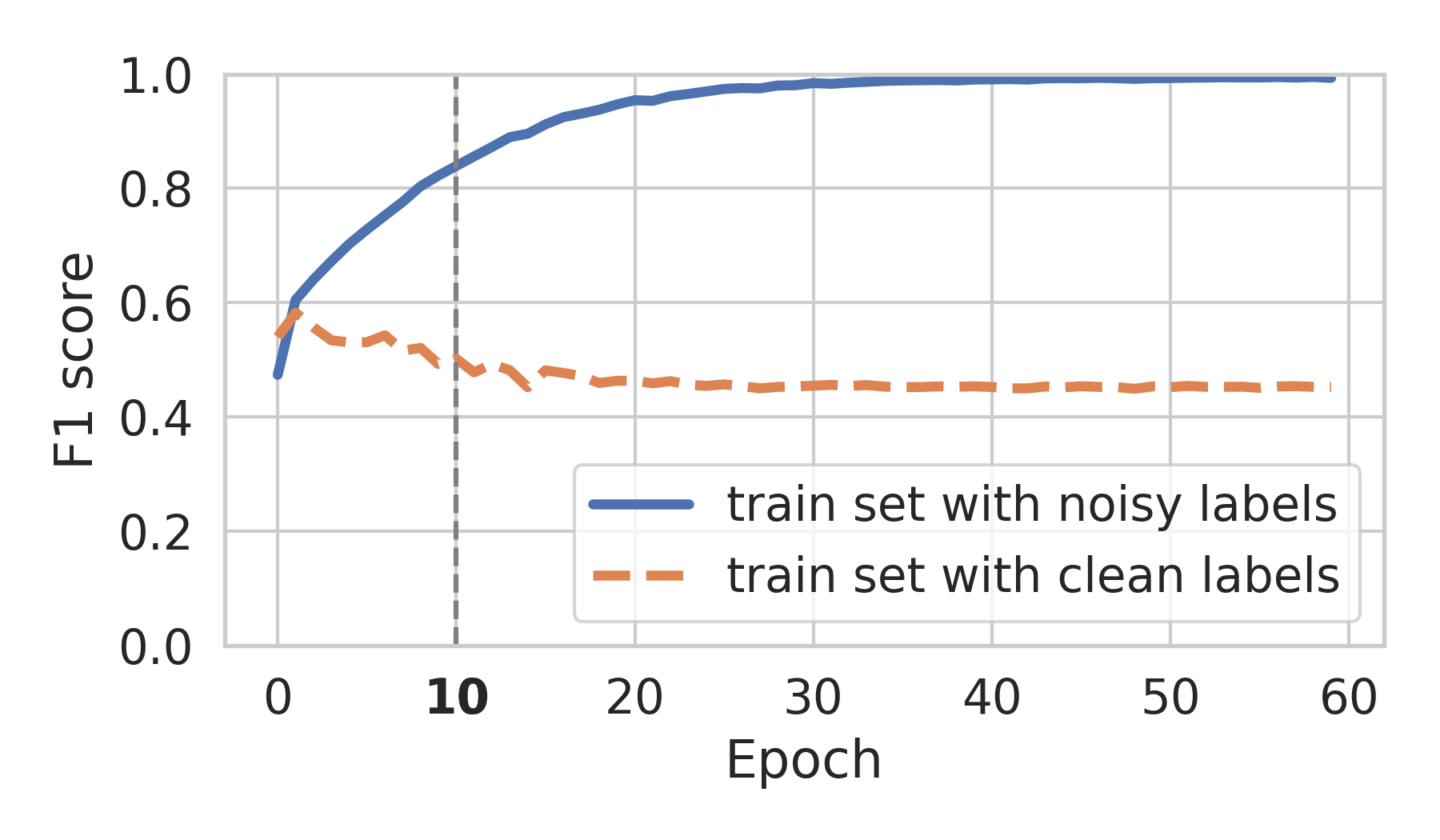} 
\caption{\noisellm~- 54.7\% noise}
\label{fig:memorization_german_llm}
\end{subfigure} 
\caption{\label{fig:memorization_german}Comparison of model performance during extended training, for the German dataset.}

\vspace{-2mm}
\end{figure}

%% file: additional_content/appendix_implementation_details.tex
\section{Implementation Details}
\label{appendix:implementation-details}

In all our experiments with noise-robust methods we use the same \texttt{xlm-roberta-large} transformer as in the FLERT baseline with a batch size of 32, except for L2R, for which we used a batch size of 16 due to VRAM constraints.

\subsection{Confident Learning}
We use regular transformer fine-tuning and obtain predicted probabilities for each sample in the training dataset 
using cross-validation. The number of folds is the only parameter in this approach and we performed a small search before choosing 3 folds. 
We use the implementation by \citet{klie2023annotation} to adapt this approach for NER by aggregating token-level predictions. We perform the final sample selection on the sentence level, training the model only using sentences that do not contain entities flagged as errors or have missing entities. 

\subsection{Co-Regularization}

We perform a hyperparameter sweep as suggested by the authors, and choose the best performing ones on the validation set of the respective noise type.

\subsection{BOND}

In our experiments we found that limiting BOND's first stage of training to 1 epoch is not enough for optimal performance, hence why we rely on the findings reported by \citet{tanzer-etal-2022-memorisation} and stop the first stage after the first 3 epochs. The second stage is limited to 7 epochs in order to reproduce the same training length as in the FLERT baseline. We update the teacher model in the second stage of training every 2 epochs as suggested by the authors for the CoNLL-03 dataset, and use hard pseudo-labels in the second stage, which we found to outperform soft pseudo-labels in our experiments.

\subsection{CrossWeigh}

We ran the CrossWeigh framework with 5 folds and 3 iterations, because according to the ablation experiments ran by the authors, higher numbers did not bring significant performance improvements. For a fair comparison, we adjust the CrossWeigh framework to use transformer fine-tuning as a base model. For the final training run using the sample weights, we used the same FLERT approach as in the baseline. 

\subsection{L2R}

We rely on the implementation provided by \citet{zhu-etal-2023-weaker}, test out two meta-learning rates while keeping the model learning rate fixed at 5e-6, and perform the validation step every 0.1 epoch with a patience of 10 validation steps. 

\subsection{MSR}

We used the implementation provided by the authors \cite{zhu-etal-2023-meta} and the hyperparameters they selected for CoNLL, as stated in their paper. For the German dataset, we used \texttt{xlm-roberta-base} as a multilingual model.

\subsection{Upper Bound: Additional Clean Data}

This upper bound assumes an additional small dataset with high-quality labels is available. We fix this number to 100 sentences, which are randomly chosen from the validation split (otherwise not used to train the models). This training setting first fine-tunes the baseline model for 10 epochs, and then continues fine-tuning only on the small clean dataset for 5 more epochs.

\subsection{GPT3.5}

To obtain an LLM-annotated variant of the training splits, for \benchmark{} we used \texttt{gpt-3.5-turbo-0613}, while for German we used \texttt{gpt-3.5-turbo-0125}.

%\todo[inline]{add more details about task and class descriptions in GPT labels}

% Reviewer: The LLM noisy dataset is surprisingly noisy. From the paper, it is unfortunately unclear what “task description” and “class descriptions” they used. I would imagine that using some reasonable descriptions (such as those one would give to a human annotator) should results in an error rate below their “45.6%” (although admittedly I did not test this, so I might be wrong here).

% Answer: Various prior works have found that it is non-trivial to generate NER annotations with LLMs. We used the Fabricator framework in our experiments which found similar error rates. See Table 5 in Golde et al. (2023). Such low scores were also found by Ye et al. (2023) in a comprehensive GPT3 and GP3.5 analysis: See Figures 4 and 5, and Table 9 in their paper (linked below). Of course, research in this field is ongoing and it is plausible that generation accuracy will improve markedly. When this happens, we could add another layer of improved LLM annotations to NoiseBench. Since our benchmark is open source, this would be easy to add in a future version. In any case, we will add more information and error examples to the paper Appendix for the final version

%% file: additional_content/appendix_clean_subsets_size.tex
\section{Size of Oracle Subsets}
\label{appendix:oracle_subset_size}

In Table \ref{tab:clean_subset}, the number of clean sentences in the oracle subset for each noisy variant is shown. These oracle subsets are used as performance upper bounds, as explained in Section \ref{clean_subset}.

\begin{table}[!htp]
\small
\setlength{\tabcolsep}{4pt}
\centering
\begin{tabular}{l r r}
\toprule
 & \% of all sentences & Oracle subset size \tabularnewline \midrule
\multicolumn{2}{l}{\textit{NoiseBench split}} & \tabularnewline
\noiseclean & 100.0 & 4879 \\
\noiseexpert & 92.6 & 4483 \\
\noisecrowdbest & 79.4 & 3786 \\
\noisecrowd & 55.3 & 2554 \\
\noisedistant & 59.7 & 2728 \\
\noiseweak & 49.6 & 2294 \\
\noisellm & 38.2 & 1705  \tabularnewline \midrule
\multicolumn{2}{l}{\textit{German split}} & \tabularnewline
\noiseclean & 100.0 & 10824 \\
\noiseexpert & 81.6 & 8827 \\ 
\noisellm & 37.8 & 4095 \\ \bottomrule
\end{tabular}
\caption{\label{tab:clean_subset} Details about the oracle subset used as an upper performance bound. The table shows the percentage of clean sentences and the absolute number, for each noise type. }
\end{table}

%% file: additional_content/appendix_uniform_noise.tex
\section{Baseline for Uniform Noise}
\label{sec:appendix-simulated-noise-extended}

Table \ref{conll_baselines_real_uniform} shows the results from Experiment 1 for uniform noise. We can see the the model is quite robust to uniform noise  and that it results in higher test performance, when compared to real noise of the same level. The average difference in F1 scores in 17 percentage points, which is why we focus on the more realistic oracle class-dependent noise simulation method in the main results of Experiments 1 and 2.

\input{tables/Exp1_uniform_simulated_noise}

%% file: tables/Exp1_uniform_simulated_noise.tex
\begin{table}[htb]
\setlength{\tabcolsep}{4pt}
\centering
\small
\begin{tabular}{lccccc}
\toprule
%  &  & \multicolumn{2}{c}{ F1 }\\ \midrule
% \noiseclean &  & \multicolumn{2}{c}{93.99 ± 0.04}\\ 
  &  \multicolumn{2}{c}{ Real noise } & \multicolumn{2}{c}{ Uniform noise } & $\Delta$ \\ \cmidrule(lr){2-3} \cmidrule(lr){4-5} \cmidrule(lr){6-6} & \textit{\%Noise} & \textit{F1} & \textit{\%Noise} & \textit{F1} & \textit{F1} \\
  \midrule
\noiseclean & 0 & 94.0 \scriptsize{±0.0} & - & - & - \\
\noiseexpert & 5.5 & 89.8\scriptsize{±0.2} & 5.4 & 93.8\scriptsize{±0.3} & 4,0 \\
\noisecrowdbest & 15.3 & 86.7\scriptsize{±0.3} & 16.1 & 92.8 \scriptsize{±0.5} & 6.1 \\
\noisecrowd & 36.6 & 70.5\scriptsize{±0.6} & 36.7 & 88.4 \scriptsize{±0.3} & 17.9 \\
\noisedistant & 31.3 & 70.8\scriptsize{±0.1} & 31.7 & 90.0 \scriptsize{±0.5} & 19.3 \\
\noiseweak & 40.4 & 65.9\scriptsize{±0.4} & 42.2 & 91.7 \scriptsize{±0.2} & 25.8 \\
\noisellm & 45.6 & 62.6\scriptsize{±0.4} & 47.3 & 89.7 \scriptsize{±0.6} & 27.1 \\
Average &  & 74.4\scriptsize{±0.3} &  & 91.4\scriptsize{0.4} & 17.0 \\
\bottomrule
\end{tabular}
\caption{\label{conll_baselines_real_uniform} F1 scores on the \noiseclean~Test split of the baseline FLERT approach, fine-tuned on different noisy variants of the training set. The scores are averages of 3 runs. The column  $\Delta$ (difference) refers to the difference in F1 score on the test split when training on a dataset with real noise compared to uniform noise.
} %The scores are averaged over 3 runs. Training on the clean training set has a score of 93.99}%\scriptsize{±0.04}}
\vspace{-2mm}
\end{table}

%% file: additional_content/appendix_memorization_additional_plots.tex
\section{Memorization of \noisecrowd, \noiseweak~and \noisellm~Noise}
\label{sec:appendix-memorization}

Figure \ref{fig:memorization_figure_appendix} is an extension of Section \ref{memorization} and shows the memorization plots from Experiment 2 for the  \noisecrowd, \noiseweak~and \noisellm~dataset variants. We again observe immediate memorization of real noise and delayed memorization of simulated noise.

\input{figures/memorization_plots_additional}

%% file: figures/memorization_plots_additional.tex
\renewcommand{\tablename}{Figure}
\setcounter{table}{4} 
\begin{table*}[!htp]
\centering
\begin{tabular}{m{0.1cm} >{\centering\arraybackslash}m{4.6cm} >{\centering\arraybackslash} m{4.6cm} >{\centering\arraybackslash}m{4.6cm}}
 & \hspace{1cm}\noisecrowd & \hspace{1cm}\noiseweak & \hspace{1cm}\noisellm \\ 
\rotatebox[origin=c]{90}{Real}\vspace{1.3cm}
& \includegraphics[width=0.31\textwidth]{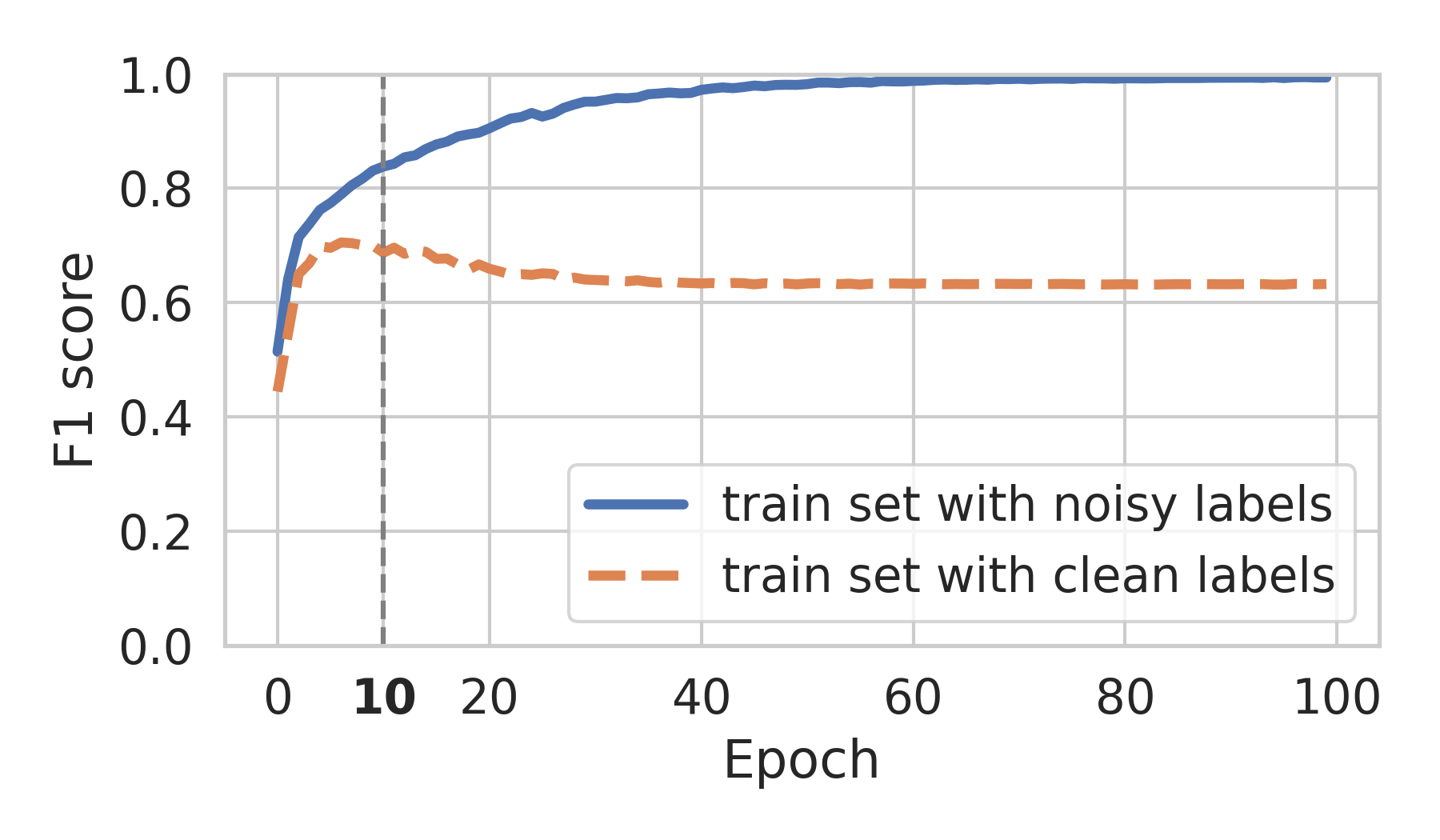} 
\subcaption{Real \noisecrowd~ - 36.6\% noise}
& \includegraphics[width=0.31\textwidth]{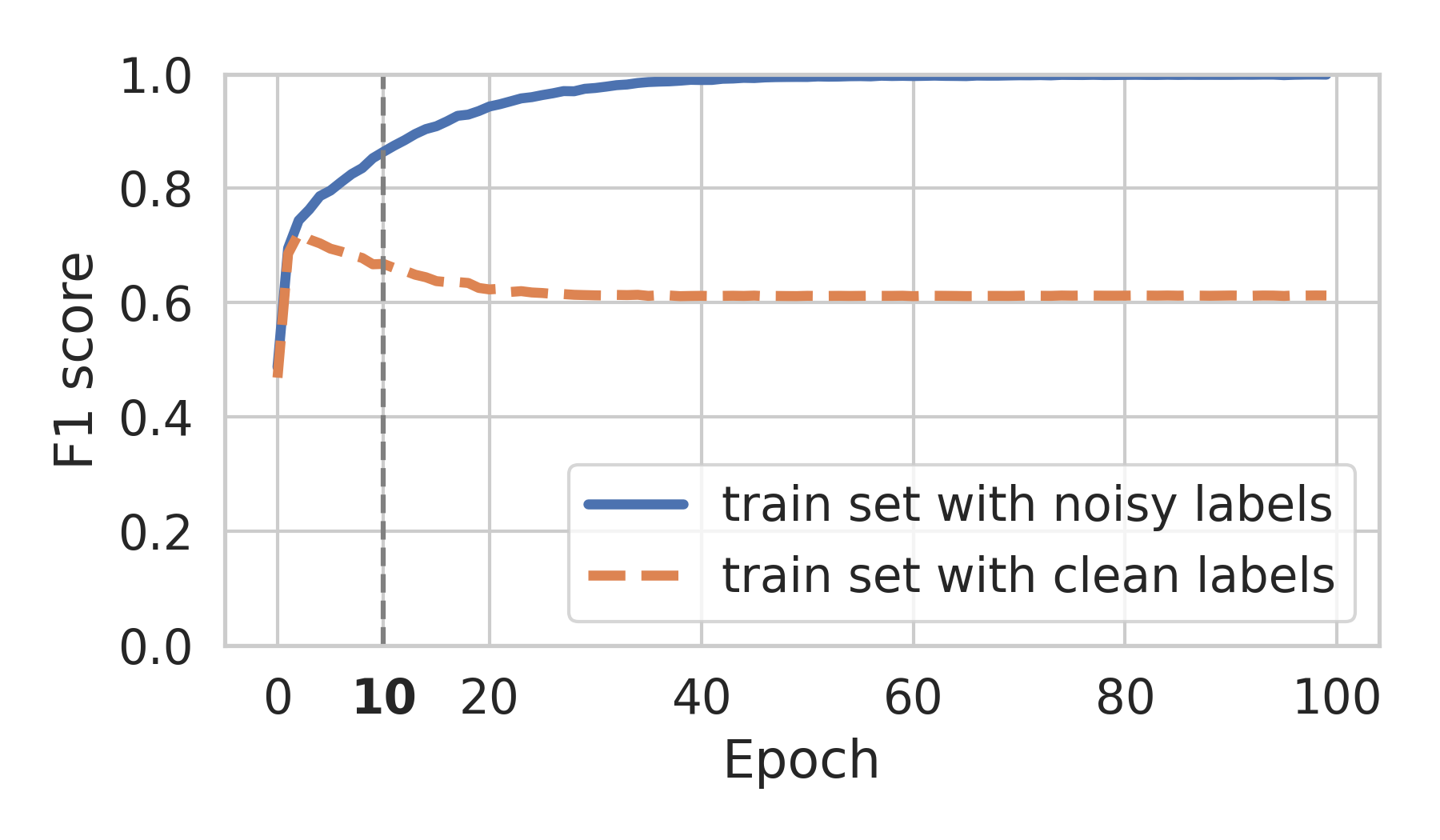} 
\subcaption{Real \noiseweak~ - 40.4\% noise}
& \includegraphics[width=0.31\textwidth]{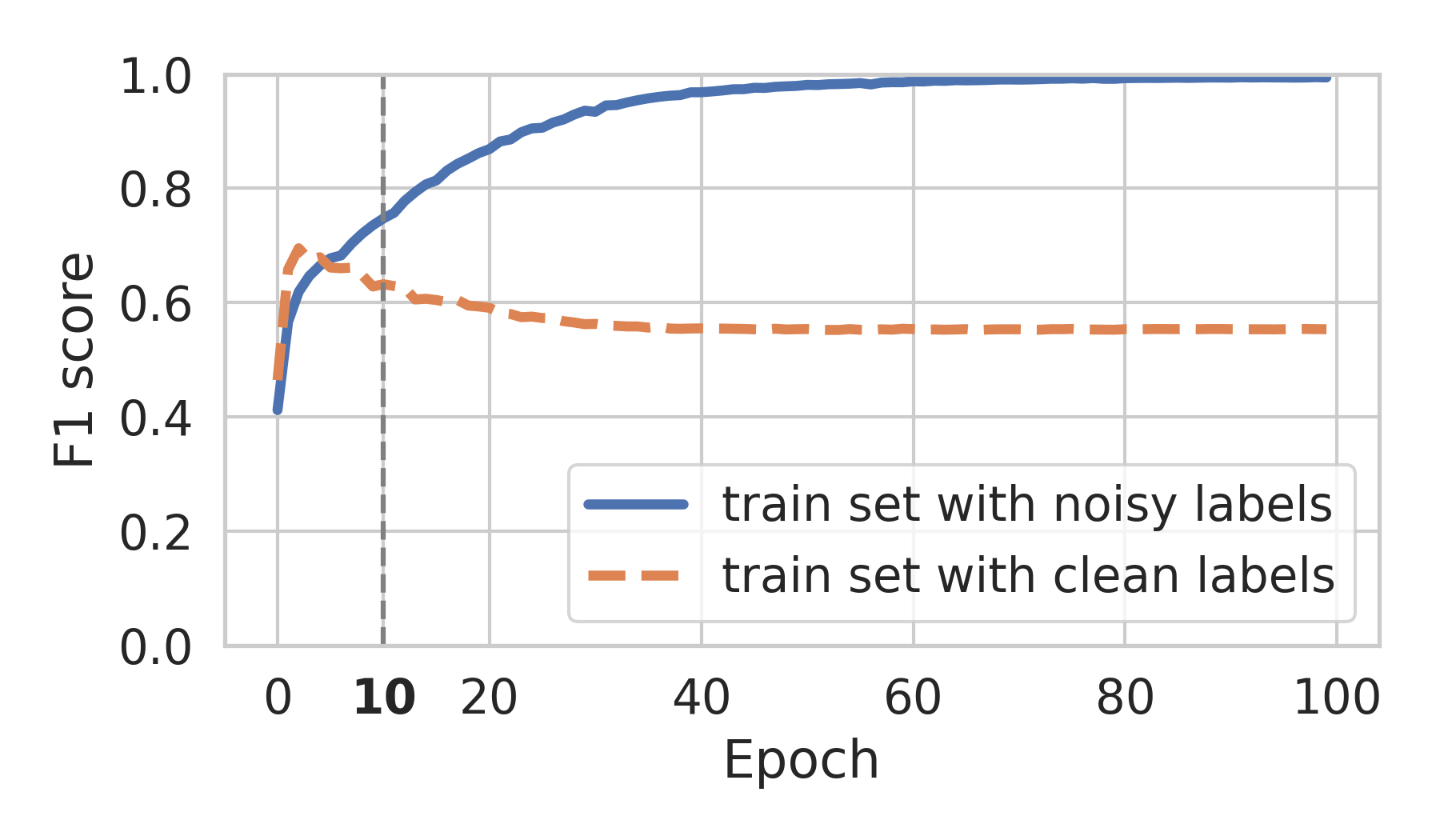}     	
\subcaption{Real \noisellm~ - 44.6\% noise} \\ 
\rotatebox[origin=c]{90}{Simulated} \vspace{1.3cm}
&  \includegraphics[width=0.31\textwidth]{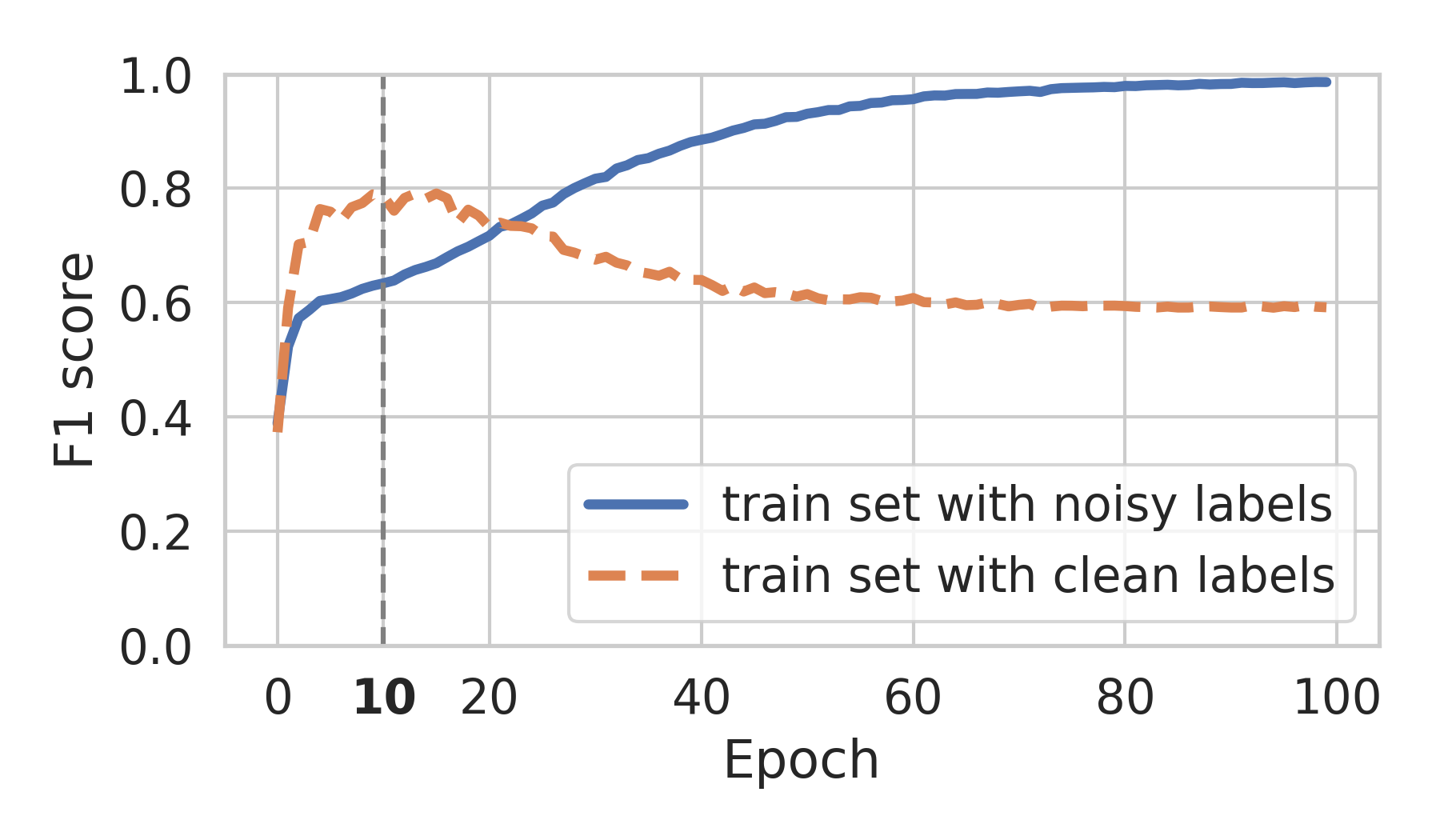} 
\subcaption{Simul. \noisecrowd~ - 41.3\% noise}
& \includegraphics[width=0.31\textwidth]{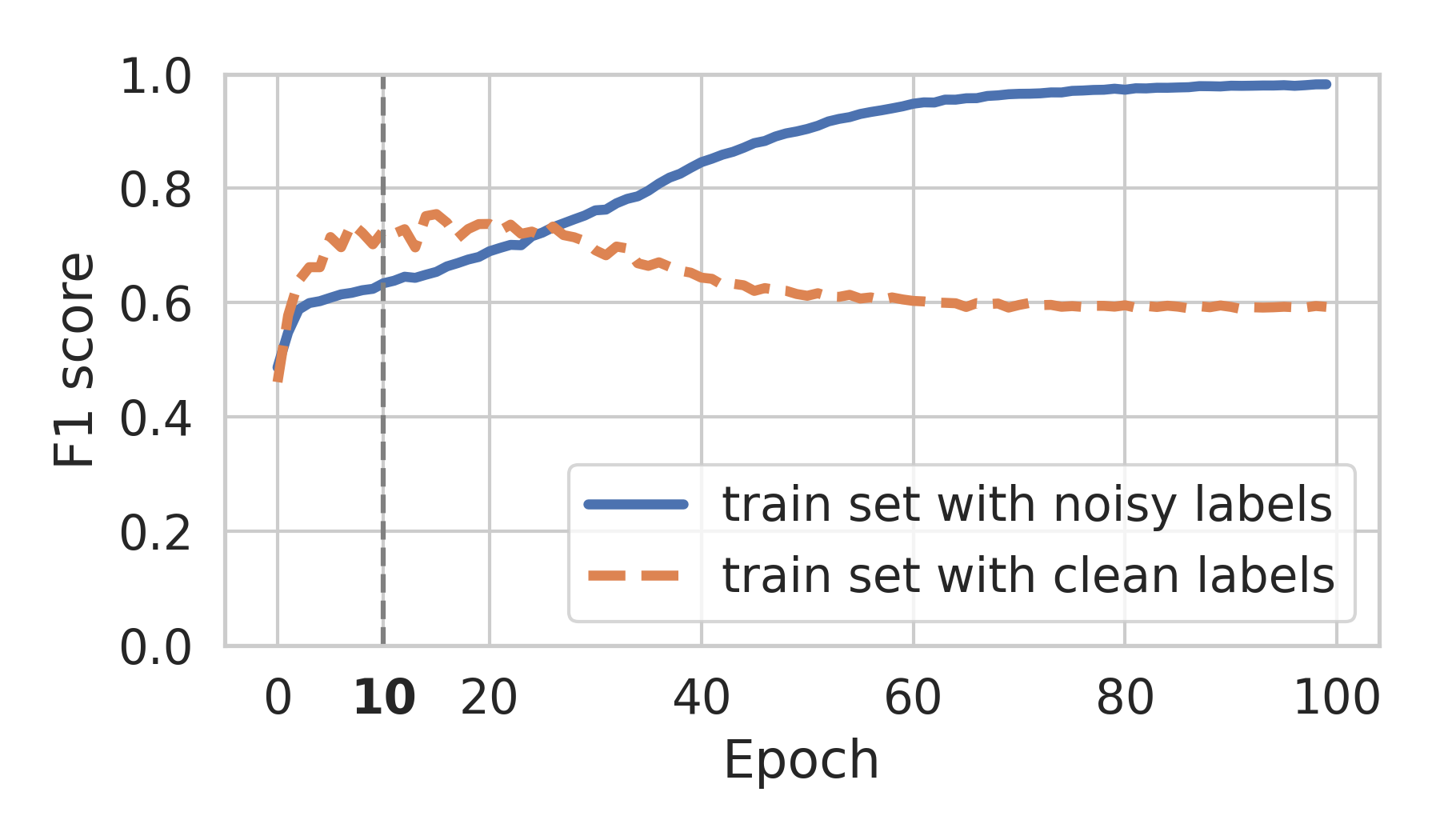} 
\subcaption{Simul. \noiseweak~ - 41.2\% noise}
& \includegraphics[width=0.31\textwidth]{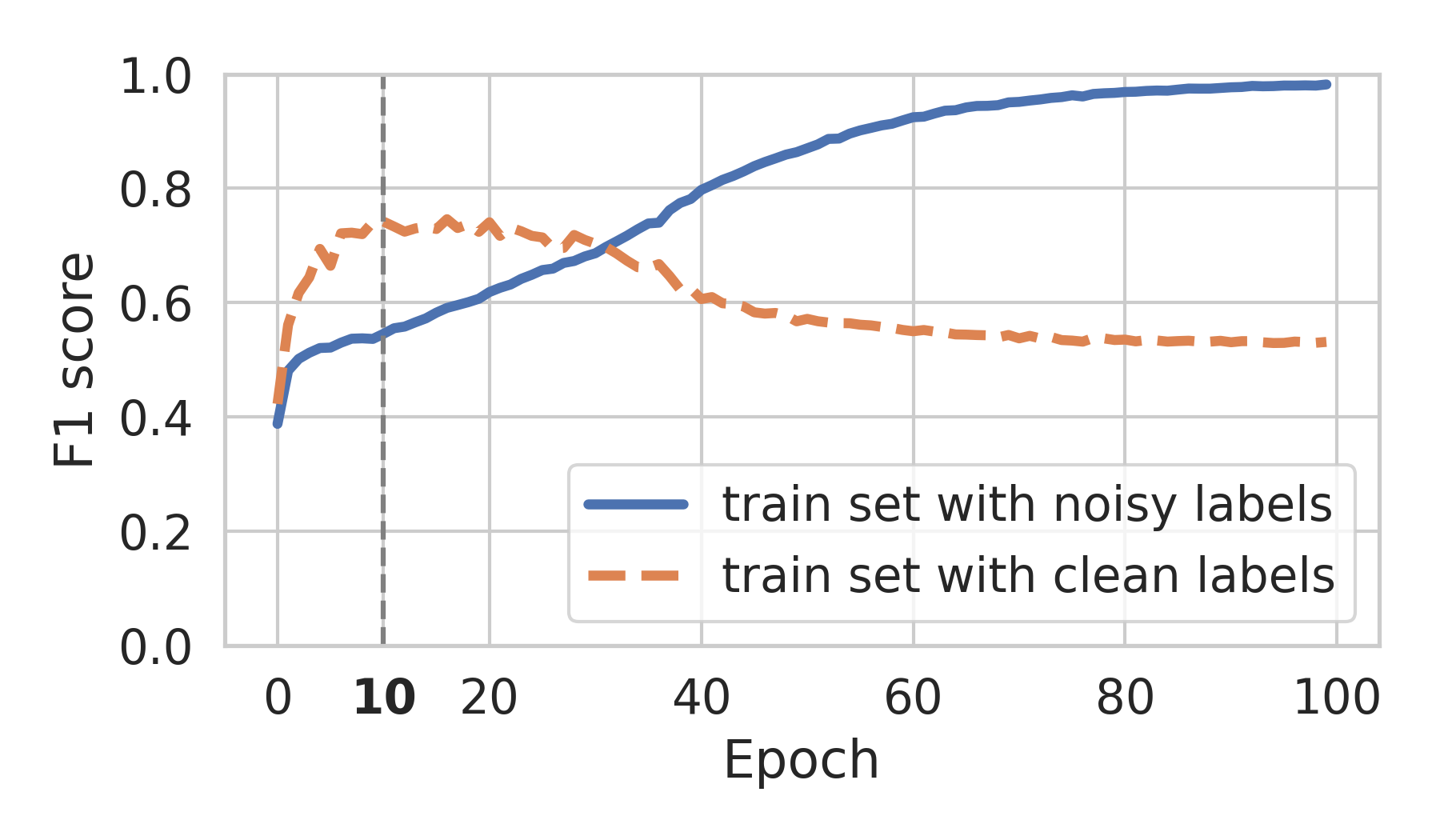} 
\subcaption{Simul. \noisellm~ - 47.2\% noise}
\\ 
\end{tabular}
\vspace{-2mm}
\caption{Comparison of model performance during extended training, for \noisecrowd, \noiseweak~and \noisellm~from \benchmark. The top row shows models fine-tuned on label sets with real noise, while the bottom row models fine-tuned on a corresponding simulated noisy label set. The graphs show both the F1 score on the noisy training labels and on the clean training labels, for 3 different noise types. The plots are averages of 3 runs.}
\label{fig:memorization_figure_appendix}
\vspace{-2mm}
\end{table*} 
\renewcommand{\tablename}{Table}
\setcounter{table}{7} 
\setcounter{figure}{5} 

%% file: additional_content/appendix_more_experiments_memorization.tex
\section{Additional Experiments on Memorization}
In addition to the main Experiment 2, we ran two ablation experiments regarding memorization. 

\input{additional_content/appendix_memorization_pretrained_vs_from_scratch}
\input{additional_content/appendix_memorization_distilroberta}

%% file: additional_content/appendix_memorization_pretrained_vs_from_scratch.tex
\subsection{Effect of Pre-training on Memorization}

The first ablation compares fine-tuning a pretrained model and a model with randomly initialized weights. Figure \ref{fig:memorization_pretraining_plot} shows this comparison during an extended training run for the \noisecrowdbest~training variant, where we used DistilBERT (learning rate of 5e-05). We can see that even without pretraining, the model starts overfitting to the noisy labels and we can observe a large gap between the performance on the clean and noisy labels.

\input{figures/memorization_grid_pretraining}

%% file: figures/memorization_grid_pretraining.tex
\begin{figure}[!htp]
\centering
\begin{subfigure}[!htb]{0.4\textwidth}
\centering
 \includegraphics[width=\linewidth]{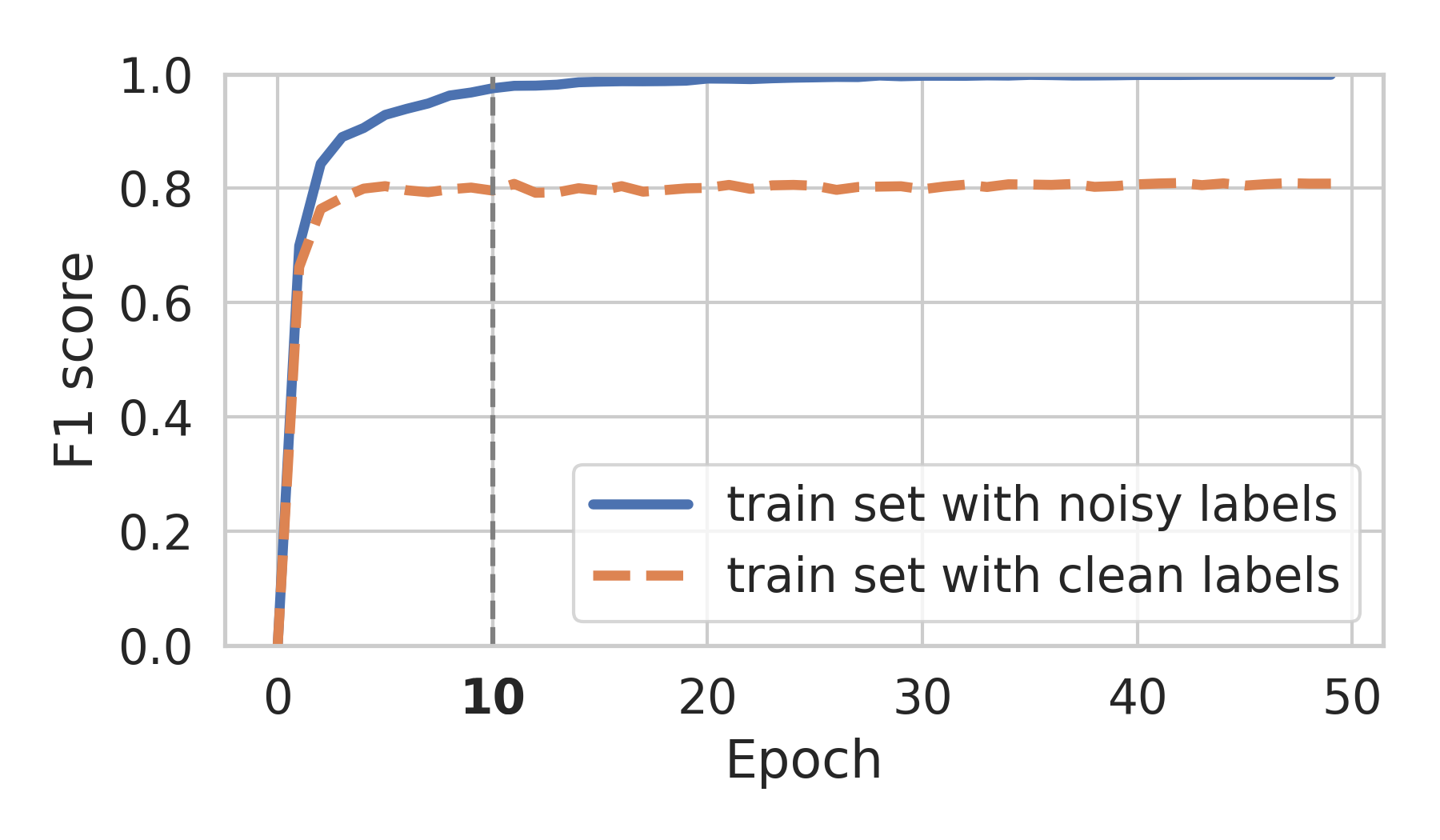} 
\caption{Pre-trained model}
\label{fig:memorization_pretraining}
\end{subfigure} 
\begin{subfigure}[!htb]{0.4\textwidth}
\centering
 \includegraphics[width=\linewidth]{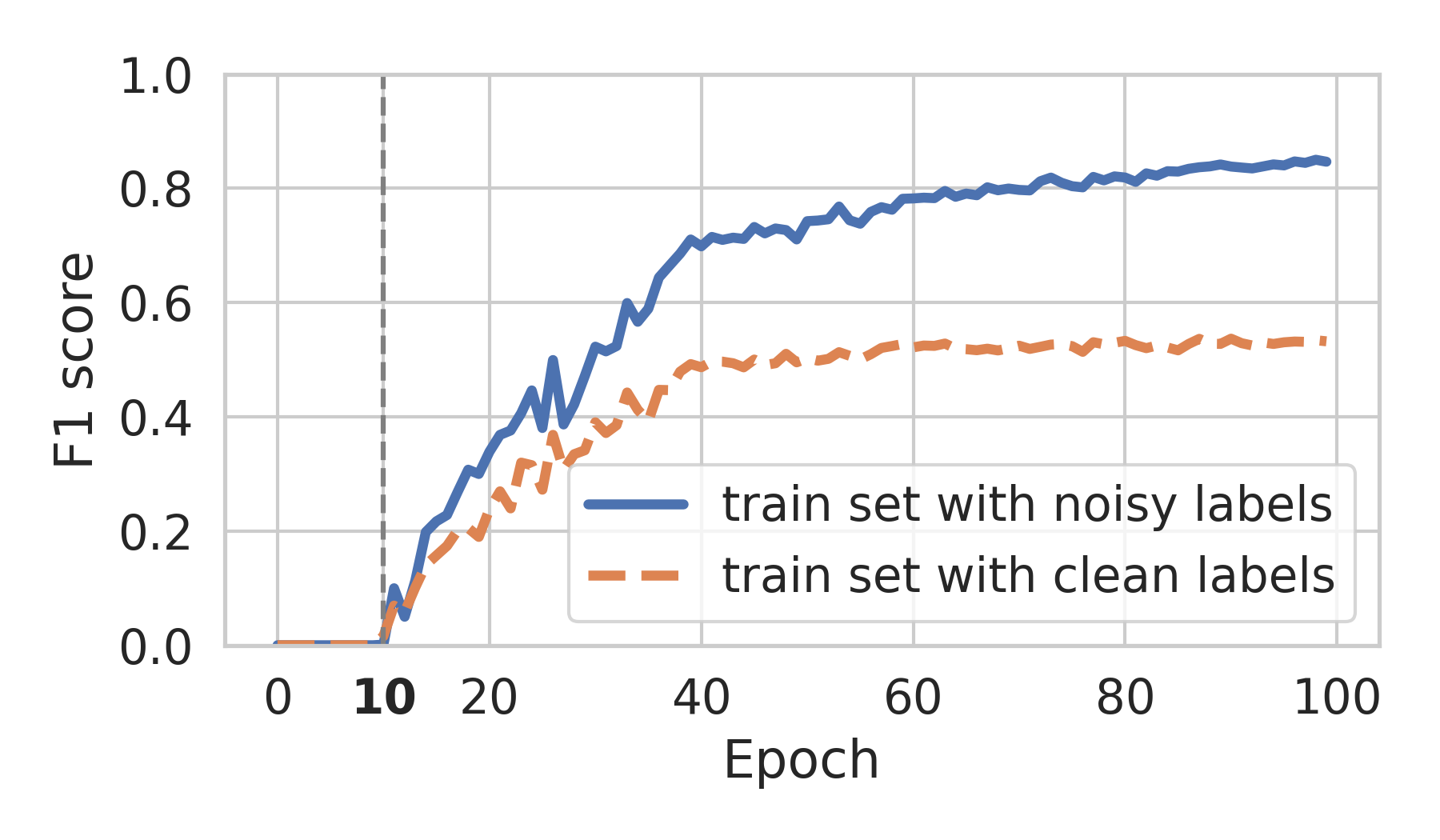} 
\caption{Random initialization}
\label{fig:memorization_no_pretraining}
\end{subfigure} 
\caption{\label{fig:memorization_pretraining_plot} Memorization of label noise in DistilBert, using the pretrained model and a model with randomly initialized weights. The experiment was run for one noise type - \noisecrowdbest.}
\vspace{-2mm}
\end{figure} 

%% file: additional_content/appendix_memorization_distilroberta.tex
\subsection{Memorization in a Smaller Model}

The second experiment investigates memorization when fine-tuning a smaller model, DistilRoBERTa, because the model we use in the main experiments, XLM-RoBERTa-Large, is quite large. Figure \ref{fig:memorization_figure_distilroberta} shows an extended training run for three noise types in NoiseBench. We observe the same patterns of immediate and delayed memorization as with the larger model.

\input{figures/memorization_plots_distilroberta}

%% file: figures/memorization_plots_distilroberta.tex
\renewcommand{\tablename}{Figure}
\setcounter{table}{6} 
\begin{table*}[!htp]
\centering
\begin{tabular}{m{0.1cm} >{\centering\arraybackslash}m{4.6cm} >{\centering\arraybackslash} m{4.6cm} >{\centering\arraybackslash}m{4.6cm}}
 & \hspace{1cm}\noiseexpert & \hspace{1cm}\noisecrowdbest & \hspace{1cm}\noisedistant \\ 
\rotatebox[origin=c]{90}{Real}\vspace{1.3cm}
& \includegraphics[width=0.31\textwidth]{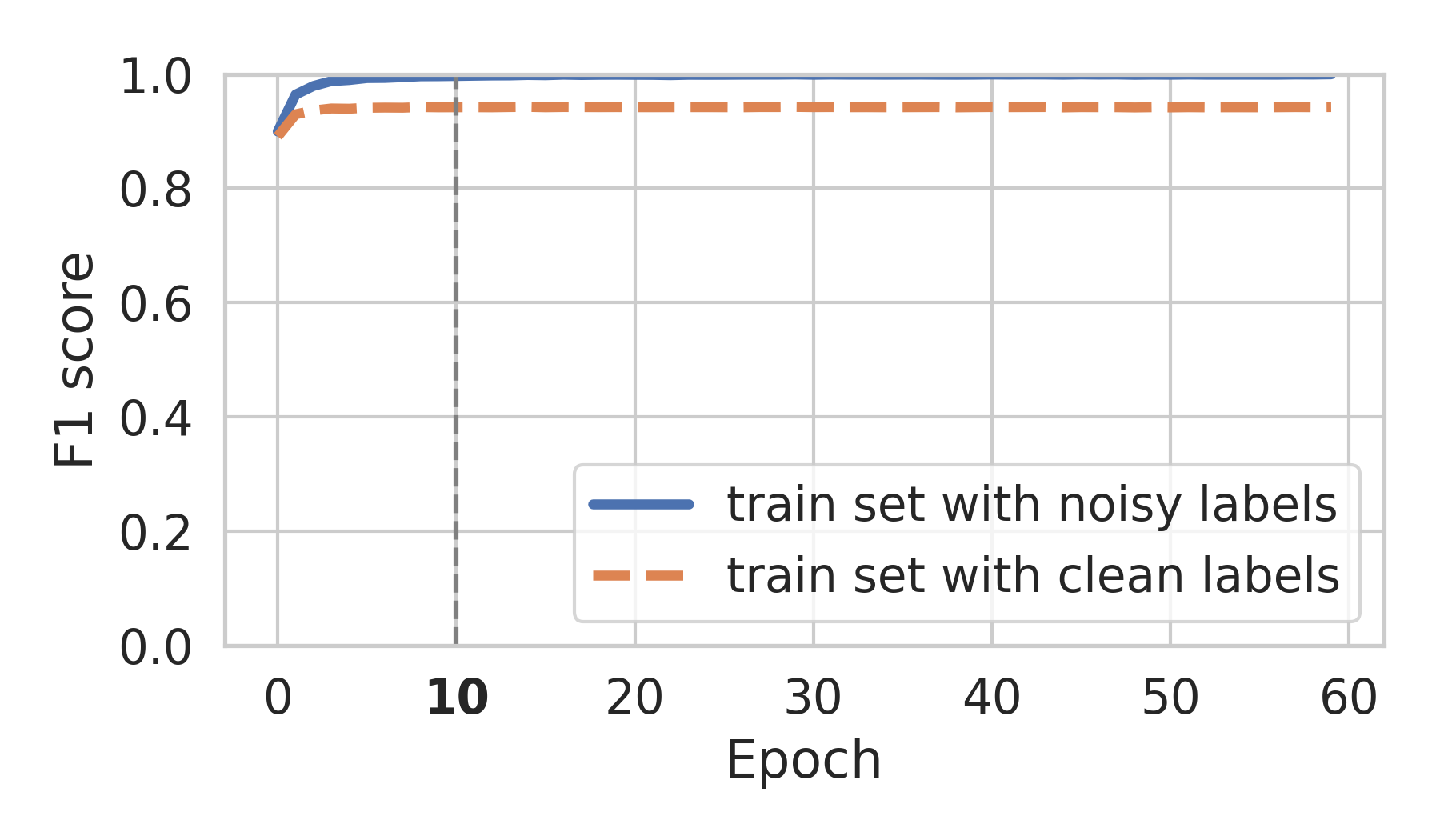} 
\subcaption{Real \noiseexpert~ - 5.5\% noise}
& \includegraphics[width=0.31\textwidth]{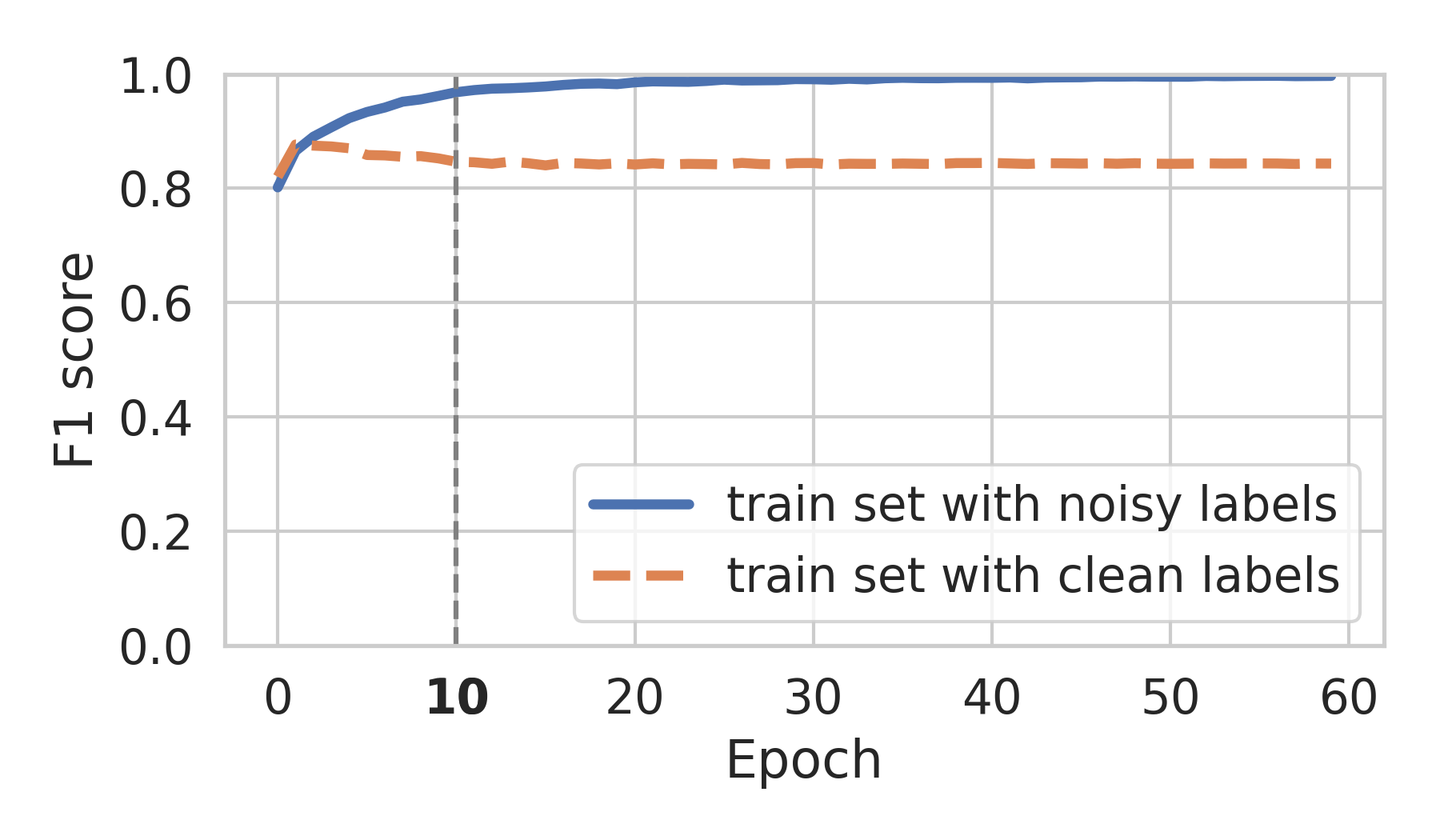} 
\subcaption{Real \noisecrowdbest~ - 18\% noise}
& \includegraphics[width=0.31\textwidth]{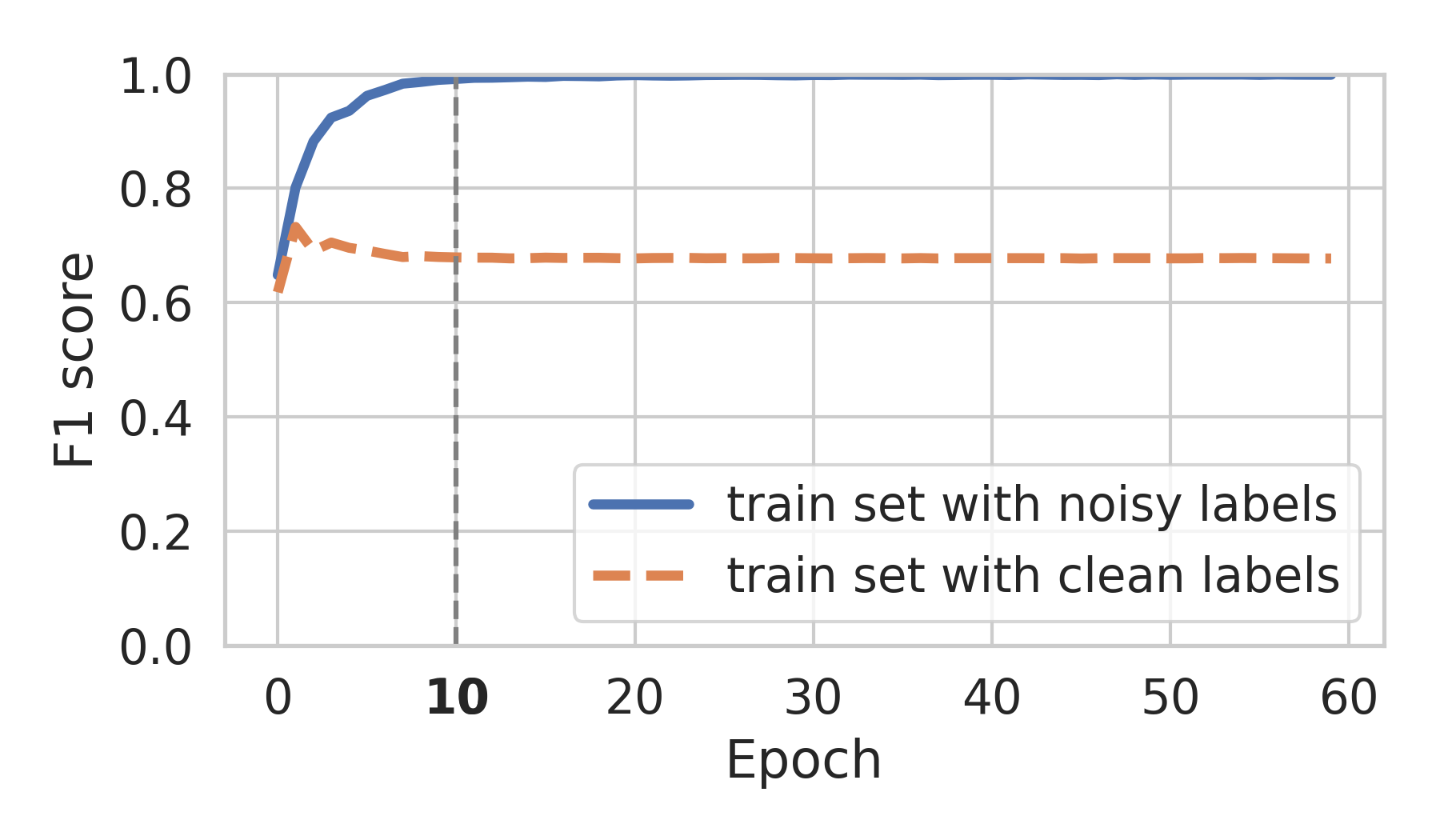}     	
\subcaption{Real \noisedistant~ - 31.3\% noise} \\ 
\rotatebox[origin=c]{90}{Simulated} \vspace{1.3cm}
&  \includegraphics[width=0.31\textwidth]{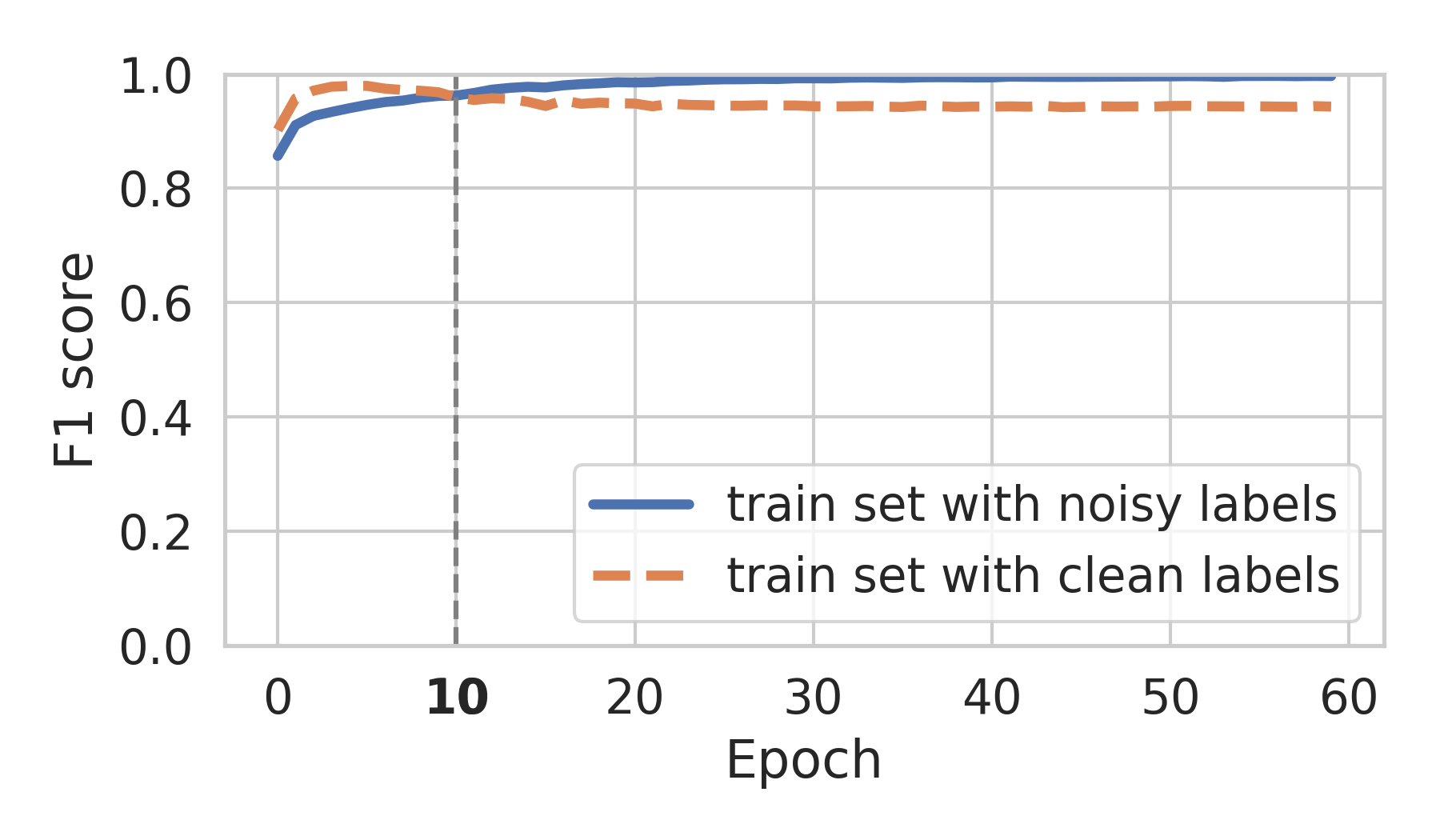} 
\subcaption{Simul. \noiseexpert~ - 6\% noise}
& \includegraphics[width=0.31\textwidth]{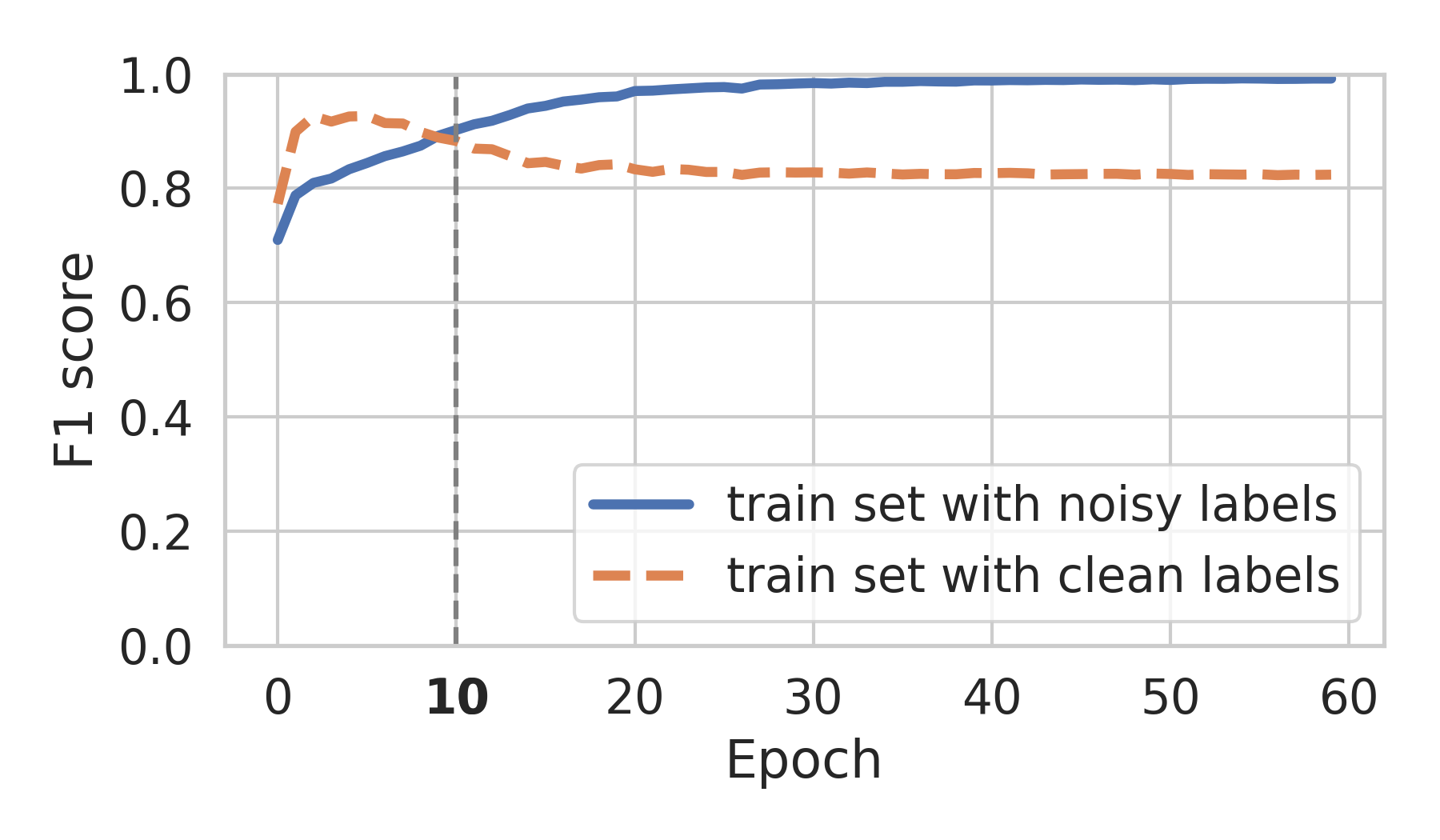} 
\subcaption{Simul. \noisecrowdbest~ - 15\% noise}
& \includegraphics[width=0.31\textwidth]{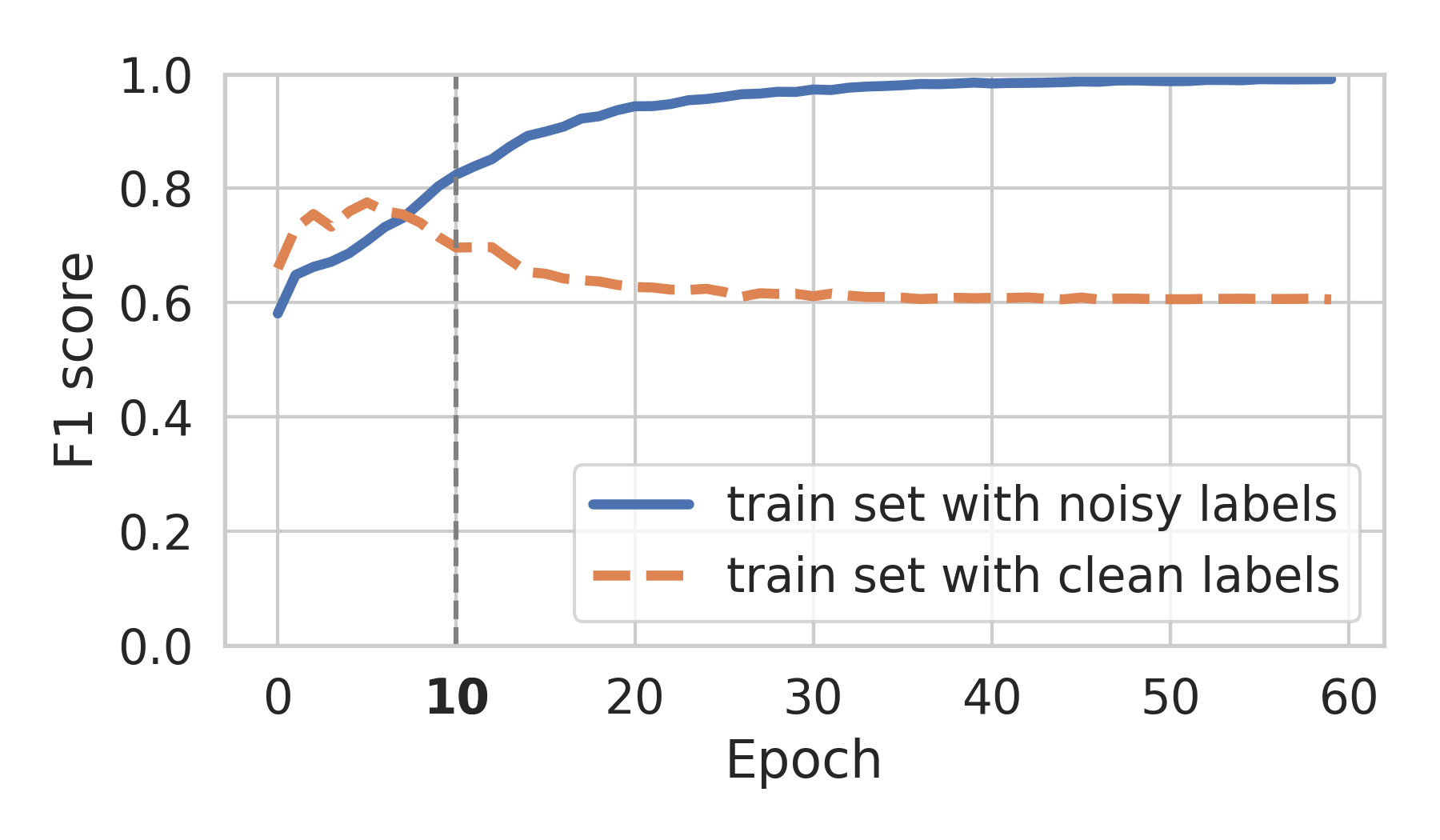} 
\subcaption{Simul. \noisedistant~ - 39\% noise}
\\ 
\end{tabular}
\vspace{-2mm}
\caption{\label{fig:memorization_figure_distilroberta}Comparison of model performance during extended training with a smaller model, Distil-RoBERTa. The top row shows models fine-tuned on label sets with real noise, while the bottom row models fine-tuned on a corresponding simulated noisy label set. The graphs show the F1 score on the noisy training labels and on the clean training labels, for 3 noise types. The plots are averages of 3 runs. }

\end{table*} 
\renewcommand{\tablename}{Table}
\setcounter{table}{7} 
\setcounter{figure}{7} 

%% file: additional_content/appendix_analysis_of_predictions_baseline_camera_ready.tex
\section{Extended Performance Metrics}
\label{sec:appendix-test-errors}

In this section we provide extended metrics of the predictive performance of the baseline FLERT method. These metrics and analysis correspond to Experiment 1 from Section \ref{Exp1}.

\input{additional_content/appendix_test_error_types}

\input{additional_content/appendix_per_class_metrics}

%% file: additional_content/appendix_test_error_types.tex
\subsection{Analysis of Test Errors}

We can characterize the model predictions in a similar way as we characterized the different types of errors in Table \ref{conll_noise_shares_table} and Figure \ref{fig:example_sentence_figure}. Table \ref{tab:error_type_analysis} shows how representative different types of prediction errors are, expressed as a percentage of all errors. We can see that with Expert noise, a majority of the mistakes are wrong entity types. Furthermore, for the Crowd and Distant dataset versions, the largest number of errors is due to missing entities, while for the Weak and LLM datasets, the errors are mostly non-entities or wrong type. This is in line with the characteristics of the noisy datasets themselves, described in Table \ref{conll_noise_shares_table}.

For German we make similar observations. For the clean variant, most errors are missing entities. However, for the two noisy variants, which include a large number of noisy non-entity annotations as seen in Table \ref{conll_noise_shares_table_german}, the majority of prediction errors are also non-entity mentions.

\begin{table}
\small
\centering
\begin{tabular}{l r r r r }
\toprule
& \multicolumn{4}{c}{\textit{\% Prediction errors}} \\
 & \textit{Missing} & \textit{Non-entity} & \textit{Type} & \textit{Partial} \\
\midrule
\multicolumn{2}{l}{\textit{NoiseBench split}} & & \tabularnewline
\noiseclean & 13.9 & 25.4 & 29.6 & 31.1 \\
\noiseexpert & 12.2 & 14.4 & 54.1 & 19.3 \\
\noisecrowdbest & 27.0 & 11.9 & 42.1 & 18.9 \\
\noisecrowd & 55.2 & 6.4 & 25.3 & 13.1 \\
\noisedistant & 64.6 & 8.2 & 14.4 & 12.8 \\
\noiseweak & 14.2 & 24.5 & 49.5 & 11.8 \\
\noisellm & 20.5 & 34.9 & 38.3 & 6.4 \\
\midrule
\multicolumn{2}{l}{\textit{German split}} & & \tabularnewline
\noiseclean & 39.9 & 26.4 & 16.7 & 16.9 \\
\noiseexpert & 17.8 & 61.9 & 11.7 & 8.6 \\
\noisellm & 26.7 & 61.5 & 4.8 & 6.9 \\
\bottomrule
\end{tabular}
\caption{\label{tab:error_type_analysis} Overview of the percentage of different types of prediction errors}

\end{table}

We also examined the confusion matrices of the predictions. We were able to identify some patterns, regarding which types of errors are more prone to memorization. For most noise types in English \benchmark{}, the largest number of prediction mistakes (out of the strings previously seen with a noisy label in the training set) were missing ORG and MISC entities, as well as ORG misclassified as LOC. These mistakes were present in a large number consistently across noise types.

However, we also observed a large number of missing ORG and MISC entities in the predictions when using the clean training set, which indicates that this is an inherently difficult pattern, even when noise is not present. On another hand, the pattern of misclassifying ORG as LOC does not happen when clean data is available. Therefore we can conclude that when this type of noisy pattern is present in the training set, the models are not able to recognize it as noise and are not robust to it.

%% file: additional_content/appendix_per_class_metrics.tex
\subsection{Per-Class Metrics}

We provide per-class metrics for a more extensive evaluation of the performance of the baseline method, for both German and English, in Table \ref{tab:perclass_metrics}. 

We can see that both precision and recall for the MISC class are generally lower than the other classes. This is especially noteworthy in the Expert label set of the German split, which does not have a high noise share, but it does have very low performance on MISC. This is however expected, as most of the noisy labels in this label set are related to MISC entities.

\input{tables/perclass_metrics_baseline}

%% file: tables/perclass_metrics_baseline.tex
\begin{table*}[t]
 \centering 
 \small
 \setlength{\tabcolsep}{4pt}
\begin{tabular}{lcccccccccccc} 
\toprule 
 & \multicolumn{3}{c}{\textit{LOC}}  & \multicolumn{3}{c}{\textit{ORG}} & \multicolumn{3}{c}{\textit{PER}} & \multicolumn{3}{c}{\textit{MISC}} \\
   \cmidrule(lr){2-4}   \cmidrule(lr){5-7}    \cmidrule(lr){8-10}   \cmidrule(lr){11-13}   
 & \textit{Prec.} & \textit{Recall} & \textit{\#Ent.} & \textit{Prec.} & \textit{Recall} & \textit{\#Ent.} & \textit{Prec.} & \textit{Recall} & \textit{\#Ent.} & \textit{Prec.} & \textit{Recall} & \textit{\#Ent.}\\ 
 \midrule
 \multicolumn{2}{l}{\textit{NoiseBench split}} & & &&&&&& \\

\noiseclean & 93.9 \scriptsize{±0.6} & 93.5 \scriptsize{±0.4} & 1413 & 93.5 \scriptsize{±0.1} & 94.6 \scriptsize{±0.1} & 1909 & 99.0 \scriptsize{±0.2} & 99.1 \scriptsize{±0.1} & 1591 & 82.9 \scriptsize{±1.0} & 86.2 \scriptsize{±1.0} & 812\\ 
\noiseexpert & 81.1 \scriptsize{±0.1} & 95.5 \scriptsize{±0.6} & 1413 & 92.0 \scriptsize{±0.6} & 81.9 \scriptsize{±0.3} & 1909 & 98.4 \scriptsize{±0.7} & 99.0 \scriptsize{±0.3} & 1591 & 85.3 \scriptsize{±0.9} & 81.4 \scriptsize{±1.1} & 812\\ 
\noisecrowdbest & 75.9 \scriptsize{±0.6} & 94.4 \scriptsize{±0.4} & 1413 & 91.8 \scriptsize{±0.3} & 75.4 \scriptsize{±0.8} & 1909 & 97.6 \scriptsize{±0.5} & 97.8 \scriptsize{±0.3} & 1591 & 86.5 \scriptsize{±0.5} & 70.8 \scriptsize{±0.1} & 812\\ 
\noisecrowd & 65.5 \scriptsize{±0.4} & 90.7 \scriptsize{±1.2} & 1413 & 83.7 \scriptsize{±0.8} & 44.5 \scriptsize{±0.3} & 1909 & 93.6 \scriptsize{±1.9} & 71.7 \scriptsize{±1.2} & 1591 & 84.4 \scriptsize{±0.4} & 47.5 \scriptsize{±0.6} & 812\\ 
\noisedistant & 83.8 \scriptsize{±0.7} & 74.8 \scriptsize{±0.2} & 1413 & 85.3 \scriptsize{±1.3} & 55.9 \scriptsize{±1.0} & 1909 & 75.7 \scriptsize{±1.7} & 84.8 \scriptsize{±0.8} & 1591 & 98.6 \scriptsize{±1.4} & 13.8 \scriptsize{±3.1} & 812\\ 
\noiseweak & 52.5 \scriptsize{±0.2} & 93.1 \scriptsize{±0.1} & 1413 & 49.9 \scriptsize{±0.2} & 34.6 \scriptsize{±0.3} & 1909 & 88.5 \scriptsize{±0.3} & 88.9 \scriptsize{±1.0} & 1591 & 84.2 \scriptsize{±0.9} & 57.3 \scriptsize{±0.9} & 812\\ 
\noisellm & 52.7 \scriptsize{±0.3} & 84.6 \scriptsize{±0.5} & 1413 & 57.7 \scriptsize{±0.9} & 45.4 \scriptsize{±0.5} & 1909 & 95.4 \scriptsize{±0.6} & 98.3 \scriptsize{±0.1} & 1591 & 12.7 \scriptsize{±0.4} & 11.8 \scriptsize{±0.4} & 812\\ 
\midrule
 \multicolumn{2}{l}{\textit{German split}} & & &&&&&& \\
\noiseclean & 93.0 \scriptsize{±0.7} & 90.8 \scriptsize{±0.2} & 1051 & 80.3 \scriptsize{±0.6} & 82.3 \scriptsize{±0.3} & 584 & 96.6 \scriptsize{±0.4} & 97.1 \scriptsize{±0.2} & 1210 & 77.1 \scriptsize{±1.7} & 56.5 \scriptsize{±0.6} & 206\\ 
\noiseexpert & 88.4 \scriptsize{±0.6} & 83.3 \scriptsize{±0.9} & 1051 & 67.1 \scriptsize{±0.8} & 84.0 \scriptsize{±0.2} & 584 & 96.9 \scriptsize{±1.0} & 96.0 \scriptsize{±0.4} & 1210 & 13.1 \scriptsize{±0.3} & 40.6 \scriptsize{±0.8} & 206\\ 
\noisellm & 63.9 \scriptsize{±0.4} & 70.3 \scriptsize{±0.9} & 1051 & 37.3 \scriptsize{±0.7} & 82.8 \scriptsize{±1.4} & 584 & 66.8 \scriptsize{±0.2} & 63.7 \scriptsize{±0.6} & 1210 & 10.4 \scriptsize{±0.6} & 18.4 \scriptsize{±1.6} & 206\\ 

\bottomrule
 \end{tabular}
 \caption{\label{tab:perclass_metrics} Per-class metrics of the predictions on the \noiseclean~test set}

 \end{table*}

%% file: noisebench.bbl
\begin{thebibliography}{37}
\expandafter\ifx\csname natexlab\endcsname\relax\def\natexlab#1{#1}\fi

\bibitem[{Arpit et~al.(2017)Arpit, Jastrz{\k{e}}bski, Ballas, Krueger, Bengio,
  Kanwal, Maharaj, Fischer, Courville, Bengio et~al.}]{arpit2017closer}
Devansh Arpit, Stanis{\l}aw Jastrz{\k{e}}bski, Nicolas Ballas, David Krueger,
  Emmanuel Bengio, Maxinder~S Kanwal, Tegan Maharaj, Asja Fischer, Aaron
  Courville, Yoshua Bengio, et~al. 2017.
\newblock A closer look at memorization in deep networks.
\newblock In \emph{International conference on machine learning}, pages
  233--242. PMLR.

\bibitem[{Chong et~al.(2022)Chong, Hong, and Manning}]{chong2022detecting}
Derek Chong, Jenny Hong, and Christopher Manning. 2022.
\newblock \href {https://doi.org/10.18653/v1/2022.emnlp-main.618} {Detecting
  label errors by using pre-trained language models}.
\newblock In \emph{Proceedings of the 2022 Conference on Empirical Methods in
  Natural Language Processing}, pages 9074--9091, Abu Dhabi, United Arab
  Emirates. Association for Computational Linguistics.

\bibitem[{Fetahu et~al.(2023)Fetahu, Chen, Kar, Rokhlenko, and
  Malmasi}]{fetahu-etal-2023-multiconer}
Besnik Fetahu, Zhiyu Chen, Sudipta Kar, Oleg Rokhlenko, and Shervin Malmasi.
  2023.
\newblock \href {https://doi.org/10.18653/v1/2023.findings-emnlp.134}
  {{M}ulti{C}o{NER} v2: a large multilingual dataset for fine-grained and noisy
  named entity recognition}.
\newblock In \emph{Findings of the Association for Computational Linguistics:
  EMNLP 2023}, pages 2027--2051, Singapore. Association for Computational
  Linguistics.

\bibitem[{Frenay and Verleysen(2014)}]{frenay2014survey}
Benoit Frenay and Michel Verleysen. 2014.
\newblock \href {https://doi.org/10.1109/TNNLS.2013.2292894} {Classification in
  the presence of label noise: A survey}.
\newblock \emph{IEEE Transactions on Neural Networks and Learning Systems},
  25(5):845--869.

\bibitem[{Golde et~al.(2023)Golde, Haller, Hamborg, Risch, and
  Akbik}]{golde2023fabricator}
Jonas Golde, Patrick Haller, Felix Hamborg, Julian Risch, and Alan Akbik. 2023.
\newblock {Fabricator: An Open Source Toolkit for Generating Labeled Training
  Data with Teacher LLMs}.
\newblock In \emph{Proceedings of the 2023 Conference on Empirical Methods in
  Natural Language Processing: System Demonstrations}. Association for
  Computational Linguistics.

\bibitem[{Hedderich et~al.(2021)Hedderich, Zhu, and
  Klakow}]{hedderich2021analysing}
Michael~A Hedderich, Dawei Zhu, and Dietrich Klakow. 2021.
\newblock Analysing the noise model error for realistic noisy label data.
\newblock In \emph{Proceedings of the AAAI Conference on Artificial
  Intelligence}, volume~35, pages 7675--7684.

\bibitem[{Huang et~al.(2021)Huang, Chen, Wu, Zhao, Xie, and
  Sun}]{huang-etal-2021-named}
Xiusheng Huang, Yubo Chen, Shun Wu, Jun Zhao, Yuantao Xie, and Weijian Sun.
  2021.
\newblock \href {https://doi.org/10.18653/v1/2021.findings-acl.423} {Named
  entity recognition via noise aware training mechanism with data filter}.
\newblock In \emph{Findings of the Association for Computational Linguistics:
  ACL-IJCNLP 2021}, pages 4791--4803, Online. Association for Computational
  Linguistics.

\bibitem[{Jiang et~al.(2020)Jiang, Huang, Liu, and Yang}]{pmlr-v119-jiang20c}
Lu~Jiang, Di~Huang, Mason Liu, and Weilong Yang. 2020.
\newblock \href {https://proceedings.mlr.press/v119/jiang20c.html} {Beyond
  synthetic noise: Deep learning on controlled noisy labels}.
\newblock In \emph{Proceedings of the 37th International Conference on Machine
  Learning}, volume 119 of \emph{Proceedings of Machine Learning Research},
  pages 4804--4815. PMLR.

\bibitem[{Klie et~al.(2023)Klie, Webber, and Gurevych}]{klie2023annotation}
Jan-Christoph Klie, Bonnie Webber, and Iryna Gurevych. 2023.
\newblock \href {https://doi.org/10.1162/coli_a_00464} {{Annotation Error
  Detection: Analyzing the Past and Present for a More Coherent Future}}.
\newblock \emph{Computational Linguistics}, 49(1):157--198.

\bibitem[{Liang et~al.(2020)Liang, Yu, Jiang, Er, Wang, Zhao, and
  Zhang}]{liang2020bond}
Chen Liang, Yue Yu, Haoming Jiang, Siawpeng Er, Ruijia Wang, Tuo Zhao, and Chao
  Zhang. 2020.
\newblock Bond: Bert-assisted open-domain named entity recognition with distant
  supervision.
\newblock In \emph{ACM SIGKDD International Conference on Knowledge Discovery
  and Data Mining}.

\bibitem[{Lison et~al.(2020)Lison, Barnes, Hubin, and
  Touileb}]{lison-etal-2020-named}
Pierre Lison, Jeremy Barnes, Aliaksandr Hubin, and Samia Touileb. 2020.
\newblock \href {https://doi.org/10.18653/v1/2020.acl-main.139} {Named entity
  recognition without labelled data: A weak supervision approach}.
\newblock In \emph{Proceedings of the 58th Annual Meeting of the Association
  for Computational Linguistics}, pages 1518--1533, Online. Association for
  Computational Linguistics.

\bibitem[{Liu et~al.(2022)Liu, Xu, Xiang, Wu, He, Zhang, and
  Zhu}]{liu-etal-2022-noise}
Bo~Liu, Wandi Xu, Yuejia Xiang, Xiaojun Wu, Lejian He, Bowen Zhang, and Li~Zhu.
  2022.
\newblock \href {https://aclanthology.org/2022.coling-1.402} {Noise learning
  for text classification: A benchmark}.
\newblock In \emph{Proceedings of the 29th International Conference on
  Computational Linguistics}, pages 4557--4567, Gyeongju, Republic of Korea.
  International Committee on Computational Linguistics.

\bibitem[{Mayhew et~al.(2019)Mayhew, Chaturvedi, Tsai, and
  Roth}]{mayhew-etal-2019-named}
Stephen Mayhew, Snigdha Chaturvedi, Chen-Tse Tsai, and Dan Roth. 2019.
\newblock \href {https://doi.org/10.18653/v1/K19-1060} {Named entity
  recognition with partially annotated training data}.
\newblock In \emph{Proceedings of the 23rd Conference on Computational Natural
  Language Learning (CoNLL)}, pages 645--655, Hong Kong, China. Association for
  Computational Linguistics.

\bibitem[{Mintz et~al.(2009)Mintz, Bills, Snow, and
  Jurafsky}]{mintz2009distant}
Mike Mintz, Steven Bills, Rion Snow, and Dan Jurafsky. 2009.
\newblock Distant supervision for relation extraction without labeled data.
\newblock In \emph{Proceedings of the Joint Conference of the 47th Annual
  Meeting of the ACL and the 4th International Joint Conference on Natural
  Language Processing of the AFNLP}, pages 1003--1011.

\bibitem[{Northcutt et~al.(2021{\natexlab{a}})Northcutt, Jiang, and
  Chuang}]{northcutt2021confident}
Curtis Northcutt, Lu~Jiang, and Isaac Chuang. 2021{\natexlab{a}}.
\newblock \href {https://doi.org/10.1613/jair.1.12125} {Confident learning:
  Estimating uncertainty in dataset labels}.
\newblock \emph{J. Artif. Int. Res.}, 70:1373–1411.

\bibitem[{Northcutt et~al.(2021{\natexlab{b}})Northcutt, Athalye, and
  Mueller}]{northcutt2021pervasive}
Curtis~G Northcutt, Anish Athalye, and Jonas Mueller. 2021{\natexlab{b}}.
\newblock \href {https://openreview.net/forum?id=XccDXrDNLek} {Pervasive label
  errors in test sets destabilize machine learning benchmarks}.
\newblock In \emph{Thirty-fifth Conference on Neural Information Processing
  Systems Datasets and Benchmarks Track (Round 1)}.

\bibitem[{Reiss et~al.(2020)Reiss, Xu, Cutler, Muthuraman, and
  Eichenberger}]{Reiss2020IdentifyingIL}
Frederick Reiss, Hong Xu, Bryan Cutler, Karthik Muthuraman, and Zachary
  Eichenberger. 2020.
\newblock \href {https://api.semanticscholar.org/CorpusID:226283626}
  {Identifying incorrect labels in the conll-2003 corpus}.
\newblock In \emph{Conference on Computational Natural Language Learning}.

\bibitem[{Ren et~al.(2018)Ren, Zeng, Yang, and Urtasun}]{ren18l2rw}
Mengye Ren, Wenyuan Zeng, Bin Yang, and Raquel Urtasun. 2018.
\newblock {Learning to Reweight Examples for Robust Deep Learning}.
\newblock In \emph{ICML}.

\bibitem[{Rodrigues et~al.(2014)Rodrigues, Pereira, and
  Ribeiro}]{rodrigues2014sequence}
Filipe Rodrigues, Francisco Pereira, and Bernardete Ribeiro. 2014.
\newblock Sequence labeling with multiple annotators.
\newblock \emph{Machine learning}, 95:165--181.

\bibitem[{Rolnick et~al.(2017)Rolnick, Veit, Belongie, and
  Shavit}]{rolnick2017deep}
David Rolnick, Andreas Veit, Serge Belongie, and Nir Shavit. 2017.
\newblock Deep learning is robust to massive label noise.
\newblock \emph{arXiv preprint arXiv:1705.10694}.

\bibitem[{Rücker and Akbik(2023)}]{ruecker2023clean}
Susanna Rücker and Alan Akbik. 2023.
\newblock \href {https://alanakbik.github.io/papers/CleanCoNLL-14.pdf}
  {{CleanCoNLL: A Nearly Noise-Free Named Entity Recognition Dataset}}.
\newblock In \emph{Proceedings of the 2023 Conference on Empirical Methods in
  Natural Language Processing}.

\bibitem[{Schweter and Akbik(2021)}]{schweter2021flert}
Stefan Schweter and Alan Akbik. 2021.
\newblock \href {http://arxiv.org/abs/2011.06993} {{FLERT: Document-Level
  Features for Named Entity Recognition}}.

\bibitem[{Song et~al.(2022)Song, Kim, Park, Shin, and Lee}]{song2022learning}
Hwanjun Song, Minseok Kim, Dongmin Park, Yooju Shin, and Jae-Gil Lee. 2022.
\newblock Learning from noisy labels with deep neural networks: A survey.
\newblock \emph{IEEE Transactions on Neural Networks and Learning Systems}.

\bibitem[{T{\"a}nzer et~al.(2022)T{\"a}nzer, Ruder, and
  Rei}]{tanzer-etal-2022-memorisation}
Michael T{\"a}nzer, Sebastian Ruder, and Marek Rei. 2022.
\newblock \href {https://doi.org/10.18653/v1/2022.acl-long.521} {Memorisation
  versus generalisation in pre-trained language models}.
\newblock In \emph{Proceedings of the 60th Annual Meeting of the Association
  for Computational Linguistics (Volume 1: Long Papers)}, pages 7564--7578,
  Dublin, Ireland. Association for Computational Linguistics.

\bibitem[{Tjong Kim~Sang and
  De~Meulder(2003)}]{tjong-kim-sang-de-meulder-2003-introduction}
Erik~F. Tjong Kim~Sang and Fien De~Meulder. 2003.
\newblock \href {https://aclanthology.org/W03-0419} {Introduction to the
  {C}o{NLL}-2003 shared task: Language-independent named entity recognition}.
\newblock In \emph{Proceedings of the Seventh Conference on Natural Language
  Learning at {HLT}-{NAACL} 2003}, pages 142--147.

\bibitem[{Wang et~al.(2022)Wang, Xiao, Dong, Feng, and Zhao}]{wang2022promix}
Haobo Wang, Ruixuan Xiao, Yiwen Dong, Lei Feng, and Junbo Zhao. 2022.
\newblock {ProMix}: combating label noise via maximizing clean sample utility.
\newblock \emph{arXiv preprint arXiv:2207.10276}.

\bibitem[{Wang et~al.(2023)Wang, Zhou, Zu, Xia, Chen, Zhang, Zheng, Ye, Zhang,
  Gui, Kang, Yang, Li, and Du}]{Wang2023InstructUIEMI}
Xiao Wang, Wei Zhou, Can Zu, Han Xia, Tianze Chen, Yuan Zhang, Rui Zheng,
  Junjie Ye, Qi~Zhang, Tao Gui, Jihua Kang, J.~Yang, Siyuan Li, and Chunsai Du.
  2023.
\newblock \href {https://api.semanticscholar.org/CorpusID:258179792}
  {Instructuie: Multi-task instruction tuning for unified information
  extraction}.
\newblock \emph{ArXiv}, abs/2304.08085.

\bibitem[{Wang et~al.(2019)Wang, Shang, Liu, Lu, Liu, and
  Han}]{Wang2019CrossWeighTN}
Zihan Wang, Jingbo Shang, Liyuan Liu, Lihao Lu, Jiacheng Liu, and Jiawei Han.
  2019.
\newblock \href {https://api.semanticscholar.org/CorpusID:202540591}
  {Crossweigh: Training named entity tagger from imperfect annotations}.
\newblock In \emph{Conference on Empirical Methods in Natural Language
  Processing}.

\bibitem[{Wu et~al.(2023)Wu, Ding, Tang, Zhang, Qin, and
  Liu}]{wu-etal-2023-noisywikihow}
Tingting Wu, Xiao Ding, Minji Tang, Hao Zhang, Bing Qin, and Ting Liu. 2023.
\newblock \href {https://doi.org/10.18653/v1/2023.findings-acl.299}
  {{N}oisywiki{H}ow: A benchmark for learning with real-world noisy labels in
  natural language processing}.
\newblock In \emph{Findings of the Association for Computational Linguistics:
  ACL 2023}, pages 4856--4873, Toronto, Canada. Association for Computational
  Linguistics.

\bibitem[{Yu et~al.(2021)Yu, Zuo, Jiang, Ren, Zhao, and Zhang}]{yu2021fine}
Yue Yu, Simiao Zuo, Haoming Jiang, Wendi Ren, Tuo Zhao, and Chao Zhang. 2021.
\newblock Fine-tuning pre-trained language model with weak supervision: A
  contrastive-regularized self-training approach.
\newblock In \emph{Proceedings of the 2021 Conference of the North American
  Chapter of the Association for Computational Linguistics: Human Language
  Technologies}, pages 1063--1077.

\bibitem[{Zaratiana et~al.(2023)Zaratiana, Tomeh, Holat, and
  Charnois}]{zaratiana2023gliner}
Urchade Zaratiana, Nadi Tomeh, Pierre Holat, and Thierry Charnois. 2023.
\newblock \href {http://arxiv.org/abs/2311.08526} {{GLiNER: Generalist Model
  for Named Entity Recognition using Bidirectional Transformer}}.

\bibitem[{Zhang et~al.(2021{\natexlab{a}})Zhang, Bengio, Hardt, Recht, and
  Vinyals}]{zhang2021generalization}
Chiyuan Zhang, Samy Bengio, Moritz Hardt, Benjamin Recht, and Oriol Vinyals.
  2021{\natexlab{a}}.
\newblock \href {https://doi.org/10.1145/3446776} {Understanding deep learning
  (still) requires rethinking generalization}.
\newblock \emph{Commun. ACM}, 64(3):107–115.

\bibitem[{Zhang et~al.(2021{\natexlab{b}})Zhang, Yu, Li, Wang, Yang, Yang, and
  Ratner}]{zhang2021wrench}
Jieyu Zhang, Yue Yu, Yinghao Li, Yujing Wang, Yaming Yang, Mao Yang, and
  Alexander Ratner. 2021{\natexlab{b}}.
\newblock \href {https://openreview.net/forum?id=Q9SKS5k8io} {{WRENCH}: A
  comprehensive benchmark for weak supervision}.
\newblock In \emph{Thirty-fifth Conference on Neural Information Processing
  Systems Datasets and Benchmarks Track}.

\bibitem[{Zhou and Chen(2021)}]{zhou2021learning}
Wenxuan Zhou and Muhao Chen. 2021.
\newblock Learning from noisy labels for entity-centric information extraction.
\newblock \emph{arXiv preprint arXiv:2104.08656}.

\bibitem[{Zhu et~al.(2022)Zhu, Hedderich, Zhai, Adelani, and
  Klakow}]{Zhu2022IsBR}
D.~Zhu, Michael~A. Hedderich, Fangzhou Zhai, David~Ifeoluwa Adelani, and
  Dietrich Klakow. 2022.
\newblock \href {https://api.semanticscholar.org/CorpusID:248266745} {{Is BERT
  Robust to Label Noise? A Study on Learning with Noisy Labels in Text
  Classification}}.
\newblock \emph{ArXiv}, abs/2204.09371.

\bibitem[{Zhu et~al.(2023{\natexlab{a}})Zhu, Shen, Hedderich, and
  Klakow}]{zhu-etal-2023-meta}
Dawei Zhu, Xiaoyu Shen, Michael Hedderich, and Dietrich Klakow.
  2023{\natexlab{a}}.
\newblock \href {https://aclanthology.org/2023.eacl-main.74} {{Meta
  Self-Refinement for Robust Learning with Weak Supervision}}.
\newblock In \emph{Proceedings of the 17th Conference of the European Chapter
  of the Association for Computational Linguistics}, pages 1043--1058,
  Dubrovnik, Croatia. Association for Computational Linguistics.

\bibitem[{Zhu et~al.(2023{\natexlab{b}})Zhu, Shen, Mosbach, Stephan, and
  Klakow}]{zhu-etal-2023-weaker}
Dawei Zhu, Xiaoyu Shen, Marius Mosbach, Andreas Stephan, and Dietrich Klakow.
  2023{\natexlab{b}}.
\newblock \href {https://doi.org/10.18653/v1/2023.acl-long.796} {Weaker than
  you think: A critical look at weakly supervised learning}.
\newblock In \emph{Proceedings of the 61st Annual Meeting of the Association
  for Computational Linguistics (Volume 1: Long Papers)}, pages 14229--14253,
  Toronto, Canada. Association for Computational Linguistics.

\end{thebibliography}
